\documentclass{article}

\usepackage[utf8]{inputenc} 
\usepackage[T1]{fontenc}    
\usepackage{hyperref}       
\usepackage{url}            
\usepackage{booktabs}       
\usepackage{amsfonts}       
\usepackage{nicefrac}       
\usepackage{microtype}      
\usepackage{amsmath}
\usepackage{amssymb}
\usepackage{algorithm}
\usepackage{algpseudocode}
\usepackage{amsmath}
\usepackage{tikz}
\usepackage{float}
\usetikzlibrary{pgfplots.groupplots}
\usepackage[numbers]{natbib}
\usepackage{amscd,amssymb,stmaryrd}
\usepackage{amsmath}
\usepackage[utf8]{inputenc}
\usepackage{amstext}
\usepackage{amsthm}
\usepackage{dsfont}
\usepackage{amsmath}
\usepackage{color}
\usepackage{mathtools}
\usepackage{cases}
\usepackage{hyperref}
\usepackage[english]{babel}
\usepackage{booktabs}
\usepackage{mathrsfs}
\usepackage{subcaption}
\usepackage{graphicx}
\usepackage{amsmath}
\usepackage{lineno}
\usepackage{pgfplots}
\pgfplotsset{compat=1.18}
\usepackage{subcaption}
\usepackage{tikz}
\usepackage[preprint]{neurips_2024}
\usepgfplotslibrary{fillbetween}
\usepackage{float}
\usetikzlibrary{pgfplots.groupplots}
\usepackage{array}
\usepackage{easyReview}

\setreviewson

\title{Conformalized-KANs: Uncertainty Quantification with Coverage Guarantees for Kolmogorov-Arnold Networks (KANs) in Scientific Machine Learning}
\author{%
  Amirhossein Mollaali\textsuperscript{*}\\
  School of Mechanical Engineering\\
  Purdue University\\
  West Lafayette, IN 47906\\
  \texttt{amollaal@purdue.edu} \\
  \And
  Christian Bolivar Moya\textsuperscript{*}\\
  Department of Mathematics\\
  Purdue University\\
  West Lafayette, IN 47906\\
  \texttt{cmoyacal@purdue.edu} \\
  \AND
  Amanda A. Howard\\
  Pacific Northwest National Laboratory\\
  Richland, WA 99354\\
  \texttt{amanda.howard@pnnl.gov}\\
  \And
  Alexander Heinlein\\
  Delft University of Technology\\
  Delft Institute of Applied Mathematics \\
  2628 CD Delft, The Netherlands \\
  \texttt{a.heinlein@tudelft.nl}
  \AND
  Panos Stinis\\
  Pacific Northwest National Laboratory\\
  Richland, WA 99354\\
  \texttt{panagiotis.stinis@pnnl.gov} \\
  \And
  Guang Lin\\
  Department of Mathematics and \\
  School of Mechanical Engineering\\
  Purdue University\\
  West Lafayette, IN 47906\\
  \texttt{guanglin@purdue.edu}
}

\begin{document}
\maketitle
\begingroup
\renewcommand{\thefootnote}{\fnsymbol{footnote}}
\footnotetext[1]{Equal contribution.}
\endgroup
\begin{abstract}
        \textcolor{black}{This paper explores uncertainty quantification (UQ) methods in the context of Kolmogorov–Arnold Networks (KANs). We apply an ensemble approach to KANs to obtain a heuristic measure of UQ, enhancing interpretability and robustness in modeling complex functions. Building on this, we introduce Conformalized-KANs, which integrate conformal prediction, a distribution-free UQ technique, with KAN ensembles to generate calibrated prediction intervals with guaranteed coverage.} Extensive numerical experiments are conducted to evaluate the effectiveness of these methods, focusing particularly on the robustness and accuracy of the prediction intervals under various hyperparameter settings. We show that the conformal KAN predictions can be applied to recent extensions of KANs, including Finite Basis KANs (FBKANs) and multifideilty KANs (MFKANs). The results demonstrate the potential of our approaches to significantly improve the reliability and applicability of KANs in scientific machine learning.
\end{abstract}
\section{Introduction} \label{sec:introduction}
Scientific machine learning (SciML) merges machine learning with scientific computing to address complex problems in fields such as physics, fluids, and biology~\cite{baker_workshop_2019,karniadakis2021physics}. While promising, the efficacy of SciML is often hampered by the extensive data requirements of traditional neural networks, limiting their use in data-sparse scientific domains. Thus, there is a critical need for innovative neural network architectures that can operate effectively with limited data and adhere to the scientific principles of the datasets, enhancing model efficiency and applicability across various scientific disciplines.

Kolmogorov-Arnold Networks (KANs)~\cite{liu2024kan,liu2024kan2} are a class of neural architectures inspired by the Kolmogorov-Arnold representation theorem~\cite{kolmogorov1957representations}, which posits that any multivariate continuous function can be represented as a superposition of continuous functions of one variable. KANs leverage this theoretical foundation to model complex, high-dimensional functions efficiently. By structuring the network to decompose high-dimensional inputs into simpler, univariate functions before recombining them, KANs can offer advantages over multilayer perceptrons (MLPs), such as discovering interpretable models. However, KANs can sometimes struggle to achieve the accuracy of MLPs~\cite{yu2024kan,zeng2024kan}. Numerous variants of KANs have appeared in the literature, including physics-informed KANs (PIKANs)~\cite{shukla2024comprehensive, wang2024kolmogorov, rigas2024adaptive, patra2024physics}, temporal KANs~\cite{genet2024tkan}, wavelet KANs~\cite{bozorgasl2024wav, seydi2024unveiling, meshir2025study}, graph KANs~\cite{kiamari2024gkan, zhang2024graphkan, de2024kolmogorov}, Chebyshev KANs (cKANs)~\cite{ss2024chebyshev, mostajeran2025scaled}, fractional KANs~\cite{aghaei2024fkan}, and deep operator KANs~\cite{abueidda2024deepokan}. KANs have been applied to time-series analysis~\cite{vaca2024kolmogorov}, fluid dynamics~\cite{kashefi2024kolmogorov,toscano2024inferring}, and computer vision~\cite{azam2024suitability,cheon2024demonstrating}, among many other applications.

In our previous work, we developed Finite Basis KANs (FBKANs)~\cite{howard2024finite} and Multi-Fidelity KANs (MFKANs)~\cite{howard2024multifidelity}. FBKANs are a domain decomposition-based network architecture that enables the parallel training of several small KANs, which are only coupled within overlapping regions, providing accurate solutions for multiscale problems. For MFKANs, we implemented a training strategy that leverages a pre-trained low-fidelity model in conjunction with a small amount of high-fidelity data to enhance high-fidelity predictions. Both FBKANs and MFKANs have demonstrated accuracy and robustness when sufficient training data is available. As shown for physics-informed neural networks (PINNs)~\cite{raissi_physics-informed_2019} in~\cite{heinlein_multifidelity_2024}, both approaches may also be combined for further improvements.

Insufficient training data often leads to inaccurate point predictions in SciML, underlining the critical need for effective uncertainty quantification (UQ)~\cite{psaros2023uncertainty}. Traditional UQ approaches for SciML have primarily relied on ensemble~\cite{yang2022scalable} and Bayesian methods~\cite{lin2023b,moya2023deeponet,zhang2024bayesian, giroux2024uncertainty, hassan2024bayesian, SAHIN2024124813}, which provide heuristic estimations of uncertainty. However, these methods do not always offer definitive guarantees on the reliability of the predictions. In response to this limitation, our previous work has advanced the application of conformal prediction~\cite{vovk2005algorithmic,romano2019conformalized,angelopoulos2021gentle} to SciML, particularly in the realm of deep operator learning~\cite{MOYA2025134418,mollaali2024conformalized}. Conformal prediction is a distribution-free UQ method that ensures coverage guarantees, thereby delivering robust prediction intervals. This approach significantly enhances prediction reliability in data-scarce scenarios, meeting the high standards required for credible and provable UQ in scientific applications.  \textcolor{black}{Building on this foundation, our objective in this paper is to combine the conformal prediction technique with KANs and their variants, including FBKANs and MFKANs, to enable uncertainty-aware predictions.}

The main contributions of our work are:
\begin{itemize}
    \item  \textcolor{black}{We integrate an ensemble approach with KANs to obtain a heuristic measure for UQ, which is compatible with KAN variants such as FBKANs and MFKANs.}
    \item \textcolor{black}{Leveraging this UQ measure, we combine conformal prediction, a supervised and distribution-free UQ method, with KAN ensembles to generate calibrated prediction intervals with user-defined coverage guarantees.}
    \item We provide extensive experiments that test the properties of these prediction intervals under various hyperparameter configurations.
\end{itemize}

The remainder of the paper is organized as follows. Section~\ref{sec:background} provides background information on KANs, FBKANs, and MFKANs. Section~\ref{subsec:ensemble} develops an ensemble method for obtaining a heuristic measure of uncertainty. We then \textcolor{black}{describe} the conformal prediction UQ method tailored for KANs, FBKANs, and MFKANs, designed to deliver reliable prediction intervals with coverage guarantees in Section~\ref{subsec:conformal-prediction}. Numerical experiments demonstrating the properties and efficacy of the conformal prediction method are presented in Section~\ref{sec:numerical-experiments}. Finally, Section~\ref{sec:conclusion} concludes the paper.

\section{Background} \label{sec:background}
This section provides an overview of KANs, FBKANs, and MFKANs. For a detailed description, we refer the reader to~\cite{howard2024finite,howard2024multifidelity}.

\subsection{Kolmogorov-Arnold Networks (KANs)} \label{subsec:kan}
KANs~\cite{liu2024kan,liu2024kan2} leverage the Kolmogorov-Arnold representation theorem~\cite{kolmogorov1957representations} to efficiently approximate multivariate functions using neural network architectures. According to this theorem, any continuous multivariate function can be represented as a finite sum of compositions involving continuous univariate functions. KANs translate this theorem into a flexible architecture designed to learn complex nonlinear mappings effectively.

Formally, given an input vector \(\mathbf{x} \in \mathbb{R}^{m_0}\), a target function \(f(\mathbf{x})\) is approximated by the nested composition:

\begin{align} \label{eq:kan}
    f(\mathbf{x}) & \approx \sum_{i_{n_l-1}=1}^{m_{n_l-1}} \varphi_{n_l-1,i_{n_l}, i_{n_l-1}} \left(\sum_{i_{n_l-2}=1}^{m_{n_l-2}} \dots \left(\sum_{i_2=1}^{m_2} \varphi_{2,i_3,i_2} \left( \sum_{i_1=1}^{m_1} \varphi_{1,i_2,i_1} \left(\sum_{i_0=1}^{m_0} \varphi_{0,i_1,i_0} (x_{i_0}) \right) \right) \right) \dots  \right), \\
                  & \triangleq \mathcal{K}(\mathbf{x}). \nonumber
\end{align}
In this formulation, \(n_l\) denotes the number of layers, \(\{m_j\}_{j=0}^{n_l}\) represents the number of neurons in each layer, and \(\varphi_{i,j,k}\) are univariate activation functions. These activation functions are defined as polynomial functions of degree \(k\), represented on a grid with \(g\) points, and consist of a weighted combination of a basis function \(b(x)\) and a B-spline:
\[
    \varphi(x) = w_b b(x) + w_s~\text{spline}(x),
\]
where
\[
    b(x) = \frac{x}{1 + e^{-x}} \quad \text{and} \quad \text{spline}(x) = \sum_{i} c_i B_i(x).
\]
Here, \(B_i(x)\) are polynomial B-splines of degree \(k\), and \(c_i\), \(w_b\), and \(w_s\) represent learnable parameters.

KANs evaluate these B-splines on a pre-defined grid. Specifically, for a one-dimensional domain \([a,b]\), the grid is partitioned into \(g_1\) intervals with grid points \(\{t_0 = a, t_1, t_2, \dots, t_{g_1} = b\}\), following the approach detailed in~\cite{liu2024kan}. Moreover, the grid extension method from~\cite{liu2024kan} enables transferring coarse-grained spline representations onto a finer grid, enhancing the network’s expressive capability and flexibility.

Next, we apply domain decomposition to obtain FBKANs~\cite{howard2024finite}. We focus on the mean squared error in the loss function, rather than the sparsity-enforcing method described in~\cite{liu2024kan}. While recent work has proposed many variations of KANs, we adopt the formulation outlined in~\cite{liu2024kan} for this study. However, the techniques developed in this paper, including the domain decomposition method we present next, can be applied to many KAN variants.
\subsection{Finite Basis KANs~(FBKANs)} \label{subsec:finite_basis_kan}
Finite basis KANs (FBKANs)~\cite{howard2024finite} are inspired by the domain decomposition approach in finite basis physics-informed neural networks (FBPINNs) developed in~\cite{moseley2023finite} and extended in~\cite{dolean2024multilevel,anderson2024elm, heinlein2024multifidelity}. In particular, FBPINNs provide a domain decomposition-based architecture for physics-informed neural networks (PINNs)~\cite{raissi_physics-informed_2019}. In FBKANs, we employ parition of unity functions $\left\lbrace \omega_j \right\rbrace_{j=1,\ldots,L}$ based on an overlapping domain decomposition, such that the function approximation in equation~\eqref{eq:kan} takes the following form:
\begin{align} \label{eq:fbkan}
    f(\mathbf{x}) & \approx \sum_{j=1}^L w_j(\mathbf{x}) \left[ \sum_{i_{n_l-1}=1}^{m_{n_l-1}} \varphi^j_{n_l-1,i_{n_l}, i_{n_l-1}} \left(\sum_{i_{n_l-2}=1}^{m_{n_l-2}} \dots \left(\sum_{i_2=1}^{m_2} \varphi^j_{2,i_3,i_2} \left( \sum_{i_1=1}^{m_1} \varphi^j_{1,i_2,i_1} \left(\sum_{i_0=1}^{m_0} \varphi^j_{0,i_1,i_0} (x_{i_0}) \right) \right) \right) \dots  \right) \right], \\
                  & \triangleq \sum_{j=1}^L w_j(\mathbf{x})\mathcal{K}^j(\mathbf{x};\theta^j). \nonumber
\end{align}
Here, \(\mathcal{K}^j(x; \theta_j)\) represents the \(j\)-th KAN with trainable parameters \(\theta_j\). 

The partition of unity functions is obtained as follows. We decompose the domain \(\Omega\) into overlapping subdomains. Each subdomain \(\Omega_j\) corresponds to the interior of the support of the function \(\omega_j\), and the set of functions \(\omega_j\) forms a partition of unity. Specifically, for \(L\) overlapping subdomains, we have
\[
    \Omega = \bigcup_{j=1}^L \Omega_j, \quad \text{supp}(w_j) = \overline{\Omega_j},~\text{and } \sum_{j=1}^L w_j =1~\text{in } \Omega.
\]
We can define the partition of unity functions in several ways. In this work, we construct them based on the expression:
\[
    w_j = \frac{\hat{w}_j}{\sum_{j=1}^L \hat{w}_j},
\]
where, in one dimension,
\begin{align*}
    \hat{w}_j
    =
    \begin{cases}
        1,                                             & L=1, \\
        [1 + \cos(\pi(x_i - \mu_{ij})/\sigma_{ij})]^2, & L>1.
    \end{cases}
\end{align*}
Here, \( \mu_{ij} \), and \( \sigma_{ij} \) represent the center and half-width of each subdomain \(j\) in each direction and \(x \in \mathbb{R}^d\). As has been investigated for FBPINNs in~\cite{dolean2024multilevel}, the width of the overlap between subdomains has an impact on the model performance; generally, the performance is improved if the width of the overlap is increased.

To train the trainable parameters \(\theta = \{\theta^j\}_{j=1}^L\) of FBKANs, we use the dataset of tuples \(\mathcal{D} = \{\textbf{x}^i, f(\textbf{x}^i)\}_{i=1}^N\) to minimize the following data-driven loss function:
\[
    \mathcal{L}(\theta) = \frac{1}{N} \sum_{i = 1}^N \left( \sum_{j=1}^L w_j(\mathbf{x}^i) \mathcal{K}^j(\mathbf{x}^i;\theta^j) - f(\mathbf{x}^i)  \right)^2.
\]
This means that we simply insert the FBKAN archicture in~\eqref{eq:fbkan} into the mean squared error loss function.

\subsection{Multi-Fidelity KANs~(MFKANs)} \label{subsec:multi_fildelity_kan}
We follow~\cite{howard2024multifidelity} and draw inspiration from composite multifidelity neural networks~\cite{meng2020composite}, which consist of three networks. The first network learns from the low-fidelity training data. The second and third networks combine to capture the correlation between the low-fidelity model’s output and the high-fidelity data, providing an accurate high-fidelity prediction. Specifically, the second network models the linear correlation, while the third network captures the nonlinear correlation. Ideally, the low- and high-fidelity datasets are highly correlated, so the linear correlation explains most of the relationship, with the nonlinear correlation acting as a small adjustment. This structure makes the nonlinear network smaller and improves training robustness, especially when limited high-fidelity data is available.

Thus, we model MFKANs using three components: a low-fidelity KAN (\(\mathcal{K}_L\)), a linear KAN (\(\mathcal{K}_l\)), and a nonlinear KAN (\(\mathcal{K}_{nl}\)). \color{black}{The low-fidelity KAN approximates the low-fidelity data. Its output, combined with the original input, serves as input to both the linear and nonlinear KANs. The linear KAN, acting as a simple linear mapping, captures linear correlations among these combined inputs and the high-fidelity data. It is implemented without any hidden layers and uses first-degree polynomial basis functions, making it suitable for learning only global linear trends. In contrast, the nonlinear KAN is specifically designed to model and capture residual nonlinearities that the linear KAN inherently cannot represent, thus refining the prediction further. This is enabled by using higher-degree polynomial basis functions and additional hidden layers in the nonlinear KAN, which increase its expressive power beyond that of the linear KAN.}


Let us consider two datasets: a low-fidelity dataset with labeled pairs \(\{(\mathbf{x}_i, f_L(\mathbf{x}_i))\}_{i=1}^{N_{LF}}\) and a high-fidelity dataset \(\{(\mathbf{x}_j, f_H(\mathbf{x}_j))\}_{j=1}^{N_{HF}}\). Following~\cite{howard2024multifidelity}, we note that \(\{\mathbf{x}_j\}_{j=1}^{N_{HF}}\) does not need to be a subset of \(\{\mathbf{x}_i\}_{i=1}^{N_{LF}}\).

\textit{Low-Fidelity KAN.} We pretrain a low-fidelity KAN and freeze its weights during high-fidelity KAN fine-tuning. To train the parameters~\(\theta_L\) of the low-fidelity KAN, we optimize the following loss function using the low-fidelity dataset:
\[
    \mathcal{L}(\theta_L) 
    = 
    \frac{1}{N_{LF}} \sum_{i=1}^{N_{LF}} \left(\mathcal{K}_L(\mathbf{x}_i;\theta_L) - f_L(\mathbf{x}_i) \right)^2.
\]

\textit{High-Fidelity KAN.} Following~\cite{howard2024multifidelity}, the high-fidelity KAN prediction is a convex combination of the linear and nonlinear KANs expressed as:
\[
    \mathcal{K}_H(\mathbf{x}) 
    = 
    \alpha \mathcal{K}_{nl}(\mathbf{x}) + (1 - \alpha) \mathcal{K}_l(\mathbf{x}).
\]
Here, \(\alpha\) is a trainable parameter. We train the high-fidelity KAN parameters \(\theta_H = \{\theta_l, \theta_{nl}\}\) using the high-fidelity dataset to minimize the following loss function:
\begin{align} \label{:eq:HF-loss}
    \mathcal{L}_H(\theta_H, \alpha) = \sum_{j=1}^{N_{HF}} \left( \mathcal{K}_H(\mathbf{x}_j; \theta_H) - f_H(x_j) \right)^2 + \lambda_\alpha \alpha^{n} + w \sum_{l=0}^{L-1} \left\| \Phi_{nl} \right\|.
\end{align}
In the above, \(n\) is a scalar hyper-parameter, typically set to \(n = 4\). This value minimizes \(\alpha\), encouraging the method to focus on learning the maximum linear correlation. However, if \(n\) is too small, \(\alpha\) becomes too small, preventing the method from capturing the full nonlinear correlation, which limits its expressiveness. If a strong nonlinear correlation exists between the high- and low-fidelity data, a larger value of \(n\) may be chosen, though this may require more high-fidelity data for training. Similarly, the hyper-parameter \(\lambda_\alpha\) is also selected before training. The final term in the above loss function represents the sum of the mean squared values of each KAN layer in the nonlinear network, inspired by~\cite{howard2024multifidelity}, and is given by
\[
    \left\| \Phi_{nl} \right\| = \frac{1}{n_{\text{in}}n_{\text{out}}}\sum_{i=1}^{n_{\text{in}}} \sum_{j=1}^{n_{\text{out}}}|\phi_{i,j}^{nl}|^2.
\]

Finally, we set \(w = 0\) or \(w = 1\) based on any known correlation between the low-fidelity and high-fidelity data. If we expect the correlation to be more strongly linear, a larger value of \(w\) forces learning a larger linear correlation. 


\section{Uncertainty Quantification Methodology} \label{sec:uq}
KANs, including FBKANs and MFKANs, provide only point predictions, which may be inaccurate. Since SciML applications demand robust solutions, we must rigorously estimate model uncertainty. In this section, we first introduce ensemble KANs to empirically measure uncertainty. Then, we apply split conformal prediction—a supervised, distribution-free UQ method—to generate rigorous prediction intervals for KANs using the ensemble predictions.

\subsection{Ensemble of KANs} \label{subsec:ensemble}

This section develops an ensemble of \textit{KAN-based networks} to enhance prediction accuracy and model robustness. To achieve this, we train an ensemble of models, each initialized randomly, to capture model uncertainty and improve generalization.

Each model in the ensemble KANs is trained by optimizing the loss function, starting from a different random initialization denoted by \(\theta_0\), where \(\theta_0\) represents the initial parameter state before training begins. After training an ensemble of \(M\) models, {\color{black} from different random initial conditions~\(\theta^j_0\)}, we obtain the parameter set \(\{\theta^j\}_{j=1}^M\). For a given test point \(x_\text{test}\), we compute ensemble statistics by performing a forward pass through all \(M\) models. Specifically, we calculate the mean and standard deviation of the \(M\)-ensemble of KANs as follows:
\begin{align*}
    \mu_{M}(x_\text{test})    & = \frac{1}{M} \sum_{j=1}^M \mathcal{K}(x_\text{test}; \theta^j),                                                \\
    \sigma_{M}(x_\text{test}) & = \sqrt{\frac{1}{M} \sum_{j=1}^M \left( \mathcal{K}(x_\text{test}; \theta^j) - \mu_M(x_\text{test}) \right)^2}.
\end{align*}
Here, \(\mathcal{K}(\cdot ; \theta^j)\) represents the forward pass of a single model in the \(M\)-ensemble, \(\mu_{M}(\cdot)\) denotes the mean prediction of the ensemble, and \(\sigma_{M}(\cdot)\) corresponds to the standard deviation.

Constructing an ensemble of FBKANs and MFKANs follows a similar procedure, but in the multi-fidelity case, we begin by freezing a single low-fidelity KAN and then train the ensemble MFKANs by optimizing the high-fidelity loss function~\eqref{:eq:HF-loss}, starting from a random initialization \(\theta_H(0)\) while incorporating the frozen pretrained low-fidelity parameters \(\theta_L\).

To estimate uncertainty, we construct a 1.96-sigma prediction interval~\cite{yang2022scalable,lin2023b} for a given test input \(x_\text{test}\). First, we compute the mean \(\mu_M(x_\text{test})\) and standard deviation \(\sigma_M(x_\text{test})\) of the ensemble predictions from the trained \(M\)-ensemble models. The 1.96-sigma prediction interval is then defined as the interval:
\[
    [\mu_M(x_\text{test}) - 1.96\sigma_M(x_\text{test}), \mu_M(x_\text{test}) + 1.96\sigma_M(x_\text{test})].
\]
This set captures the range within which the true output \(f(x_\text{test})\) is expected to lie with approximately 95\% confidence, assuming the predictions follow a normal distribution. 

In SciML, assuming that model predictions follow a normal distribution is often incorrect. Consequently, providing a 95\% coverage guarantee for ensemble predictions is inherently difficult. The uncertainty in real-world datasets, especially in those corresponding to complex scientific problems, often deviates from the assumptions of normality. To address this challenge, we turn to split conformal prediction in the next section. This method leverages the ensemble predictions to generate prediction intervals with rigorous coverage guarantees, ensuring more reliable uncertainty estimation without relying on normality assumptions.
\subsection{Conformal Prediction for KANs} \label{subsec:conformal-prediction}
Conformal prediction~\cite{vovk2005algorithmic,romano2019conformalized,angelopoulos2021gentle} offers a robust framework for creating prediction intervals with guaranteed coverage, providing a distribution-free approach to UQ in machine learning models. This method is particularly valuable for SciML applications where data distributions often defy conventional assumptions. At its core, conformal prediction ensures that a prediction interval includes the true output with a predefined probability, independent of the underlying data distribution. This attribute establishes conformal prediction as a versatile and reliable tool in scenarios where traditional statistical methods fall short.

To build these prediction intervals, conformal prediction leverages a model trained on a specific dataset to generate a heuristic measure of uncertainty. For our purposes, we use the standard deviation of the ensemble’s predictions as this measure of uncertainty.

Algorithm~\ref{alg:conformal-KAN} outlines the procedure for applying conformal prediction to ensemble KANs, including ensemble FBKANs and MFKANs. In particular, to apply conformal prediction effectively we require a \textit{calibration dataset} that must be exchangeable. This exchangeability ensures that the order of data points does not impact the prediction interval, thereby ensuring the model's conformal predictions remain consistent and reliable for new, unseen data. Notably, since i.i.d.~data inherently implies exchangeability, assuming i.i.d.~data in the calibration set guarantees that the predictions are valid across the dataset. In our configuration, the calibration dataset comprises \(n\) data pairs, \(\mathcal{D}_{\text{cal}} = \{x_j, f(x_j)\}_{j=1}^n\), where \(x_j\) is the input and \(f(x_j)\) is the corresponding output. These tuples are essential for estimating the nonconformity scores, which are crucial for generating accurate prediction intervals.

\begin{algorithm}
    \caption{Split Conformal Prediction for KAN-based model}
    \begin{algorithmic}[1]
        \State \textbf{Input:} Calibration data $\mathcal{D}_\text{cal} = \{(x_j, f(x_j))\}_{j=1}^n$, miscoverage level $\alpha$, and trained ensemble  KAN-based model \(\mu_M, \sigma_M\).
        \State \textbf{Output:} Confidence intervals for new test examples.
        \State Use ensemble  KAN-based model to predict the mean and standard deviations \(\mu_M(x_j), \sigma_M(x_j)\).
        \State Calculate nonconformity scores \(s_1, s_2, \dots, s_n\) for \(\mathcal{D}_{\text{cal}}\), where:
        \[
            s_j =\frac{|f(x_j) - \mu_M(x_j)|}{\sigma(x_j)}.
        \]
        \State Determine the empirical $\lceil(n+1)(1-\alpha)\rceil/n$-quantile of the set of scores, denote as $\hat{q}_\alpha$.
        \For{each new test point $x_{\text{test}}$}
        \State Obtain from ensemble KANs-based models \(\mu_M(x_\text{test}), \sigma_M(x_\text{test})\).
        \State Construct the prediction interval:
        \[
            \mathcal{C}_\alpha(x_{\text{test}}) = [\mu_M(x_\text{test}) - \hat{q}_\alpha \cdot \sigma_M(x_{\text{test}}), \mu_M(x_\text{test}) + \hat{q}_\alpha \cdot \sigma(x_{\text{test}})].
        \]
        \EndFor
        \State \textbf{return} Confidence intervals $\mathcal{C}_\alpha(x_{\text{test}})$ ensuring at least $1 - \alpha$ coverage.
    \end{algorithmic}
    \label{alg:conformal-KAN}
\end{algorithm}
We use the calibration dataset \(\mathcal{D}_{\text{cal}}\) to compute nonconformity scores \(s_1, \dots, s_n\). These scores evaluate the degree to which each data point in the calibration set aligns with the predictions of the trained KAN-based model, considering the inherent uncertainty. Essentially, a data point with a low nonconformity score indicates good alignment between the KAN's predictions and the actual outcomes, suggesting its inclusion in the prediction interval. In ensemble-based models like ours, we estimate this uncertainty by calculating the standard deviation of the ensemble's predictions. Therefore, for each tuple \((x_j, f(x_j))\) in \(\mathcal{D}_{\text{cal}}\), we determine the \(j\)-th nonconformity score \(s_j\) as follows~\cite{romano2019conformalized,MOYA2025134418,mollaali2024conformalized}:
\[
    s_j =\frac{|f(x_j) - \mu_M(x_j)|}{\sigma(x_j)}.
\]
To obtain the prediction interval, we compute \(\hat{q}_\alpha\) as the \(\lceil(n+1)(1-\alpha)\rceil/n\) empirical quantile of the set of nonconformity scores \(s_1, \dots, s_n\). Using this quantile \(\hat{q}_\alpha\), for a new test input \(x_\text{test}\), the proposed conformal prediction method for KANs produces a prediction interval:
\[
    \mathcal{C}_\alpha(x_{\text{test}}) = [\mu_M(x_{\text{test}}) - \hat{q}_\alpha \cdot \sigma_M(x_{\text{test}}), \mu_M(x_{\text{test}}) + \hat{q}_\alpha \cdot \sigma(x_{\text{test}})].
\]
Here, the prediction interval is ideally small and includes the target \(f(x_{\text{test}})\) with a high probability of \(1-\alpha\), as selected by the user.

\begin{figure*}[h!]
    \centering
    \includegraphics[width=0.85\textwidth]{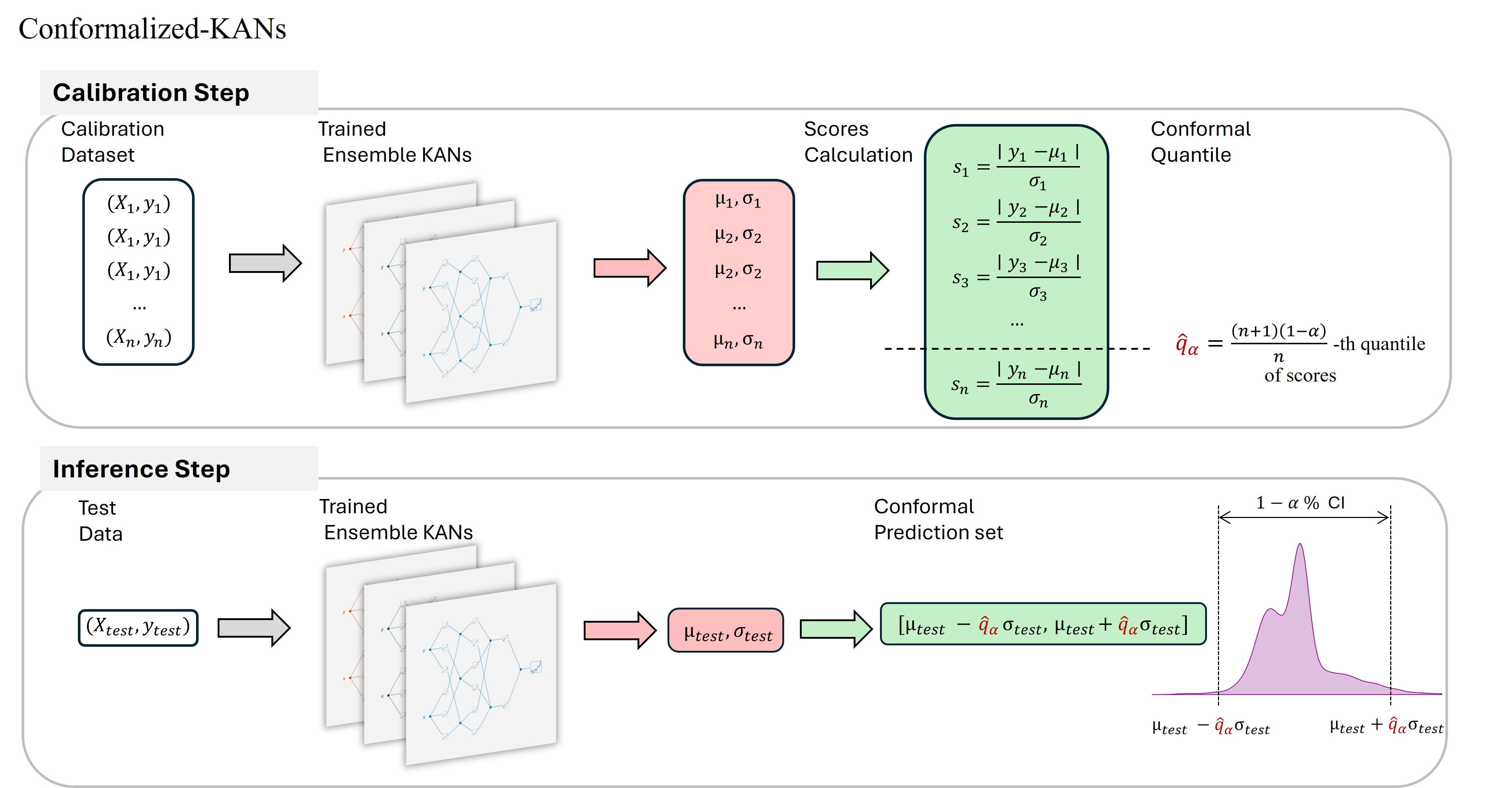}
\caption{
Schematic representation of the Conformalized-KANs framework. \textbf{Calibration step:} A dedicated calibration dataset is used to estimate the conformal quantile $\hat{q_\alpha}$ corresponding to a target miscoverage rate $\alpha$. \textbf{Inference step:} This quantile is then employed to construct prediction intervals that achieve the desired coverage.
}
    \label{fig:Confkab-Schematic}
\end{figure*}

The theoretical guarantees for the prediction intervals generated by the proposed conformal prediction method for KANs are outlined next. Given that the data is exchangeable, the method assures coverage:
\[
    \mathbb{P}(f(x_\text{test}) \in \mathcal{C}_\alpha(x_\text{test})) \ge 1 - \alpha,
\]
Note that while i.i.d. random variables are necessarily exchangeable, the converse may not hold. Therefore, our datasets in the experiments are always i.i.d.. For prediction intervals of the form \(\mathcal{C}_\alpha = \{y = f(x_\text{test}) : s(x_\text{test}, y) \le \hat{q}_\alpha\}\), including as proposed in this paper, the theorem in~\cite{vovk2005algorithmic,angelopoulos2021gentle} ensures coverage guarantees of the form:
\[
    1 - \alpha \le \mathbb{P}(f(x_\text{test}) \in \mathcal{C}_\alpha(x_\text{test})) \le 1 - \alpha + \frac{1}{n+1}.
\]
Figure \ref{fig:Confkab-Schematic} illustrates the schematic of the Conformalized-KANs framework. Within this framework, the ensemble KANs can comprise any ensemble of trained KAN models, including FBKAN and MFKAN.

\section{Numerical Experiments} \label{sec:numerical-experiments}
This section evaluates the performance of the proposed Conformalized-KANs in ensuring guaranteed coverage. The effectiveness of this approach is demonstrated through four approximation tasks: a 1-D function, a 2-D function, a multi-fidelity problem, and a PDE problem. In each case, an ensemble of KAN networks is trained to generate a heuristic measure of uncertainty and provide prediction intervals. Conformal prediction is then applied to refine these intervals, ensuring the desired coverage level. Across all experiments, the objective is to predict the 95\% confidence interval, corresponding to a miscoverage rate of \(\alpha = 0.05\).
\subsection{Metrics}
To evaluate the performance of the proposed Conformalized-KANs, we use the following metrics.

\textit{Average Coverage: }
This metric quantifies the proportion of observed values that fall within the predicted intervals, reflecting the model's reliability. It is defined as:

\[
    \text{Average Coverage} = \frac{1}{N} \sum_{i=1}^{N} I(y_i \in [\hat{y}_{i, \text{lower}}, \hat{y}_{i, \text{upper}}]),
\]

where \(N\) is the total number of data points, \textcolor{black}{\(y_i = f(x_i)\)} is the observed value, and \([\hat{y}_{i, \text{lower}}, \hat{y}_{i, \text{upper}}]\) represents the predicted interval. The indicator function \(I(\cdot)\) returns 1 if \(y_i\) falls within the interval and 0 otherwise. A higher Average Coverage indicates better reliability in uncertainty estimation.

\textit{Prediction Interval Width (PIW):} Average PIW represents the average width of predicted intervals, providing insight into model uncertainty:

\[
    \text{Average PIW} = \frac{1}{N} \sum_{i=1}^{N} |\hat{y}_{i, \text{upper}} - \hat{y}_{i, \text{lower}}|.
\]

A narrower Average PIW suggests more confident predictions, but overly tight intervals may lead to insufficient coverage. Ideally, Average PIW should be minimized while maintaining high Average Coverage to balance accuracy and UQ.

\subsection{Experiment 1: 1-D Function}
In this experiment, we consider the following 1-D function:
\[
    f(x) = \exp(\sin(0.3 \pi x^2)),
\]
on \(x \in [0,2]\). This function was selected due to its nonlinear and locally varying structure, which presents a meaningful challenge for predictive modeling. The sinusoidal component introduces periodic variations, creating regions of differing smoothness, while the exponential transformation amplifies these variations, making the function more dynamic and highlighting areas of higher uncertainty. We test this case using both KAN and FBKAN networks.

\textit{KAN Design and Datasets.}
To generate prediction intervals, we use an ensemble of size 4 for both KAN and FBKAN networks. For FBKANs, the number of subdomains is set to 10. The dataset comprises 400 random samples for training, 500 for calibration in conformal prediction, and 1,000 for evaluation.

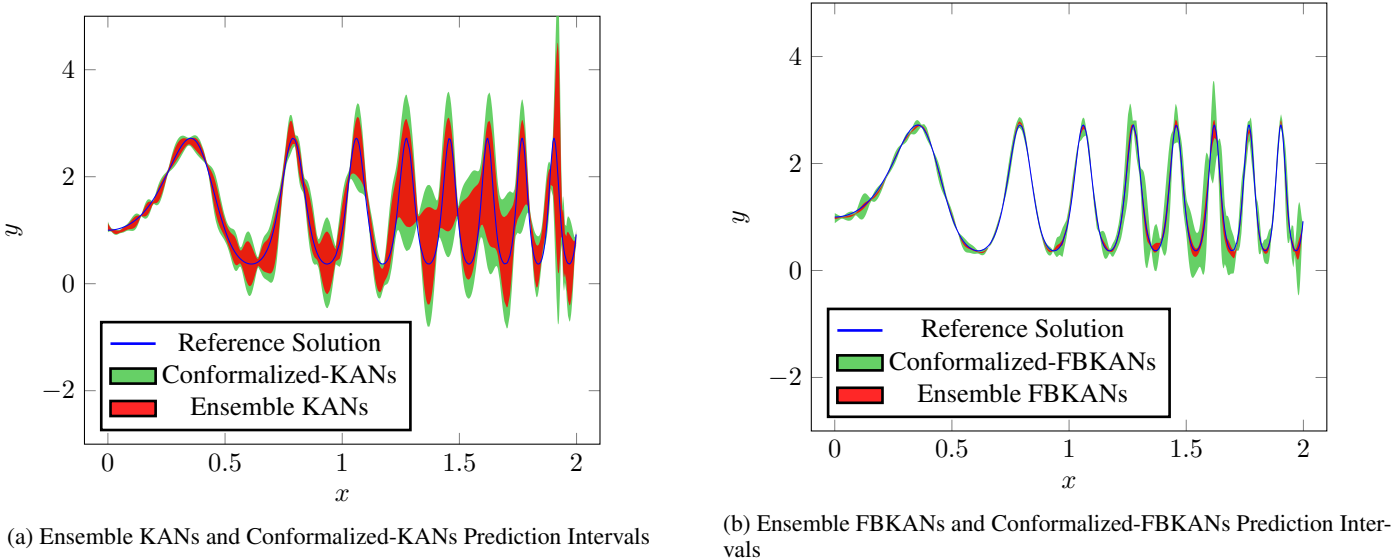
\begin{figure}[h!]
    \begin{subfigure}[!bl]{0.50\textwidth}
        \begin{tikzpicture}
            \begin{axis}[ xlabel={$x$}, ylabel={$y$},xmin=-0.1, xmax=2.1,ymin=-3, ymax=5, legend pos=south west, line width=1pt,enlargelimits=false,
                    ytick={-2, 0, ..., 6},
                ]

                \addplot[solid, color=blue]
                table{figsData/1st_experiment/y_test.txt};
                \addlegendentry{Reference Solution}

                \addplot[name path=conf_kan_low,  fill=none, draw=none,forget plot]
                table{figsData/1st_experiment/conf_kan_low.txt};
                \addplot[name path=conf_kan_high, fill=none, draw=none,forget plot]
                table{figsData/1st_experiment/conf_kan_high.txt};

                \definecolor{mycolor1}{RGB}{75,200,75}
                \addplot[fill=mycolor1, fill opacity=0.85] fill between[of=conf_kan_high and conf_kan_low, ];
                \addlegendentry{Conformalized-KANs}

                \addplot[name path=kan_low,  fill=none, draw=none,forget plot]
                table{figsData/1st_experiment/kan_low.txt};
                \addplot[name path=kan_high, fill=none, draw=none,forget plot]
                table{figsData/1st_experiment/kan_high.txt};

                \addplot[fill=red, fill opacity=0.85] fill between[of=kan_high and kan_low, ];
                \addlegendentry{Ensemble KANs}

            \end{axis}
        \end{tikzpicture}
        \caption{Ensemble KANs and Conformalized-KANs Prediction Intervals}
        \label{subfig:Prob-DeepONet}
    \end{subfigure}
    \hspace{20pt}
    \begin{subfigure}[!br]{0.5\textwidth}
        \begin{tikzpicture}
            \begin{axis}[ xlabel={$x$}, ylabel={$y$},xmin=-0.1, xmax=2.1,ymin=-3, ymax=5, legend pos=south west, line width=1pt,enlargelimits=false,
                    ytick={-2, 0, ..., 6},
                ]

                \addplot[solid, color=blue]
                table{figsData/1st_experiment/y_test.txt};
                \addlegendentry{Reference Solution}
                \addplot[name path=conf_kan_low,  fill=none, draw=none,forget plot]
                table{figsData/1st_experiment/conf_fbkan_low.txt};
                \addplot[name path=conf_kan_high, fill=none, draw=none,forget plot]
                table{figsData/1st_experiment/conf_fbkan_high.txt};

                \definecolor{mycolor1}{RGB}{75,200,75}
                \addplot[fill=mycolor1, fill opacity=0.85] fill between[of=conf_kan_high and conf_kan_low, ];

                \addlegendentry{Conformalized-FBKANs}

                \addplot[name path=fbkan_low,  fill=none, draw=none,forget plot]
                table{figsData/1st_experiment/fbkan_low.txt};
                \addplot[name path=fbkan_high, fill=none, draw=none,forget plot]
                table{figsData/1st_experiment/fbkan_high.txt};

                \addplot[fill=red, fill opacity=0.85] fill between[of=fbkan_high and fbkan_low ];
                \addlegendentry{Ensemble FBKANs}

            \end{axis}
        \end{tikzpicture}
        \caption{Ensemble FBKANs and Conformalized-FBKANs Prediction Intervals}
        \label{subfig: fig1}
    \end{subfigure}
    \begin{center}

    \end{center}

    \caption{Conformalized prediction intervals generated on 1-D function test dataset, computed with an ensemble size of 4 and a target miscoverage rate of \(\alpha = 0.05\).}
    \label{fig:1D}
\end{figure}

\textit{Results.} Figure \ref{fig:1D} shows the predicted intervals from the ensemble KANs and ensemble FBKANs, along with their conformalized prediction intervals for the test data. Table \ref{table:1D} summarizes the results for each method. Both Conformalized-KANs and Conformalized-FBKANs achieve the target 95\% coverage, while the original, non-conformal models fail to meet the desired confidence level. Consequently, the ensemble method yields smaller average width.
Moreover, the Conformalized-KANs intervals are notably wider, reflecting greater inherent uncertainty in approximations made by standard KAN architectures. Meanwhile, Conformalized-FBKANs intervals are narrower but sufficiently accurate due to FBKAN's use of domain decomposition, enabling localized and more precise approximations (see also the ablation study in Section \ref{app:ablation-1d}).

\begin{table}[h!]
    \centering
    \begin{tabular}{c   c   | c  c }
        \hline
        \textbf{Model}      & \textbf{Average Coverage} & \textbf{Average PIW} & \textbf{Std.~Dev PIW} \\

        \hline
        Conformalized-KANs   & \textbf{96.10}\%          & \textbf{1.06}        & \textbf{0.91}\        \\
        Conformalized-FBKANs & \textbf{95.30}\%          & \textbf{0.41}        & \textbf{0.42}\        \\
        \hline
        Ensemble KANs                 & 88.30\%                   & 0.73                 & 0.62                  \\
        Ensemble FBKANs               & 52.81\%                   & 0.08                 & 0.08                  \\
        \hline
    \end{tabular}
\caption{Average coverage, average interval width, and the standard deviation of interval widths for ensemble KANs, ensemble FBKANs, and their conformalized counterparts, evaluated on 1-D function test dataset. Each method targets a 95\% prediction interval, corresponding to a miscoverage rate of \(\alpha = 0.05\).}

    \label{table:1D}
\end{table}

\subsection{Experiment 2: 2-D Function}
In this experiment, we consider the following 2-D function:
\[
    f(x,y) = \sin(6 \pi x^2) \sin(8 \pi y^2),
\]
on \((x,y) \in [0,1] \times [0,1]\).

\textit{KAN Design and Datasets.} For this experiment, an ensemble of size 4 is used for KAN and FBKAN models, with the number of subdomains for FBKAN set to 4. The dataset comprises 2,000 samples for training, 1,000 for calibration, and 1000 for evaluation.

\begin{figure*}[h!]
    \centering
    \includegraphics[width=1\textwidth]{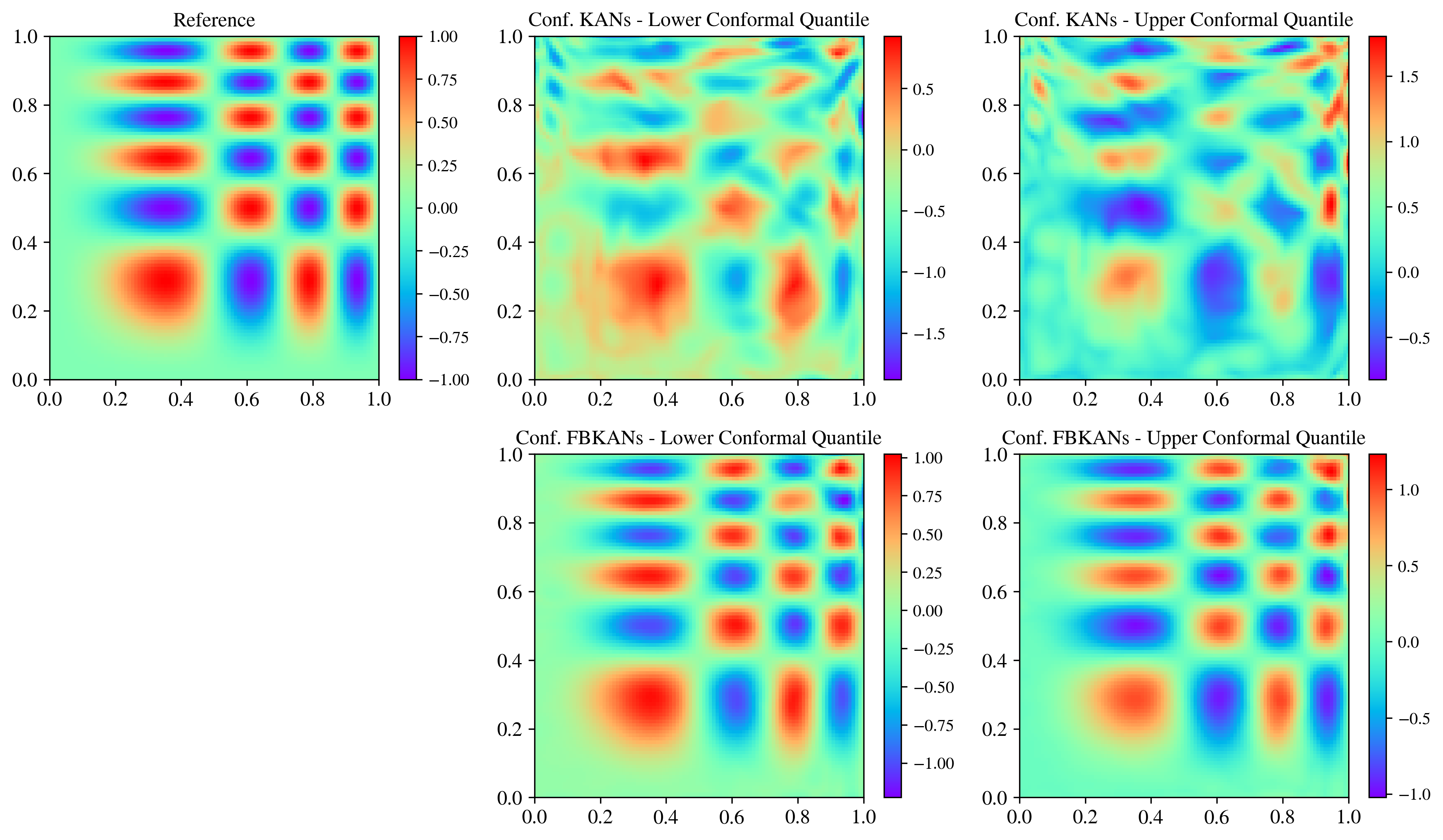}

    \caption{Conformalized prediction intervals generated by Conformalized-KANs and Conformalized-FBKANs on the 2-D function test dataset, computed with an ensemble size of 4 and a target miscoverage rate of \(\alpha = 0.05\).}
        
    \label{fig:2D-example}
\end{figure*}

\textit{Results.} Figure~\ref{fig:2D-example} presents the conformalized prediction intervals for both models, \textcolor{black}{alongside the reference solution, which corresponds to the true values of the target function \(f\) evaluated over the domain}. Table \ref{table:2D-example} summarizes the coverage and width results. As shown in the table, for the KAN network the ensemble model tends to underestimate uncertainty, while Conformalized-KANs achieves the target 95\% coverage. For the FBKAN model, the ensemble version overestimates coverage, whereas its conformalized counterpart refines the predictions, bringing them close to the target coverage level. Consequently, in this case, the Conformalized-FBKANs produces narrower prediction intervals compared to its non-conformalized counterpart.

\begin{table}[h!]
    \centering
    \begin{tabular}{c   c   | c  c }
        \hline
        \textbf{Model}      & \textbf{mean Coverage} & \textbf{Average PIW} & \textbf{Std.~Dev PIW} \\

        \hline
        Conformalized-KANs   & \textbf{95.50}\%       & \textbf{0.62}        & \textbf{0.37}\        \\
        Conformalized-FBKANs & \textbf{94.85}\%       & \textbf{0.09}        & \textbf{0.10}\        \\
        \hline
        Ensemble KANs                 & 92.44\%                & 0.49                 & 0.30                  \\
        Ensemble FBKANs               & 97.07\%                & 0.10                 & 0.12                  \\
        \hline
    \end{tabular}
    \caption{Average coverage, average interval width, and the standard deviation of interval widths for ensemble KANs, ensemble FBKANs, and their conformalized counterparts, evaluated on 2-D function test dataset. Each method targets a 95\% prediction interval, corresponding to a miscoverage rate of \(\alpha = 0.05\).}
    \label{table:2D-example}
\end{table}
\subsection{Experiment 3:  Multi-Fidelity Problem} In this experiment, we consider a multi-fidelity problem involving a jump function with a linear correlation between the low- and high-fidelity expressions. The low-fidelity function \( f_L(x) \) and the high-fidelity function \( f_H(x) \) are defined as follows:

\begin{align*}
    f_L(x) & = \begin{cases} 0.1(0.5(6x-2)^2 \sin(12x - 4) + 10(x-0.5)-5)\quad x \le 0.5, \\ 0.1(0.5(6x-2)^2 \sin(12x - 4) + 10(x-0.5)-2)\quad x > 0.5, \end{cases} \\
    f_H(x) & = 2 f_L(x) - 2x + 2,
\end{align*}

for \(x \in [0,1]\).

\textit{KAN Design and Datasets.}
To generate prediction intervals, we train an ensemble of size 5 following the procedure detailed in Section \ref{subsec:ensemble}. The dataset consists of 120 low-fidelity training samples and 5 high-fidelity samples for training. Additionally, 40 samples are used for calibration, and 200 data points are allocated for evaluating the MFKAN models.

\begin{figure}[t!]
    \centering
    \begin{tikzpicture}
        \begin{axis}[
                width=0.7*\linewidth,height =0.45*\linewidth,
                xlabel={$x$}, ylabel={$y$},xmin=-0.1, xmax=1.1,ymin=-0.5, ymax=2.5, legend pos=north west, line width=1pt,enlargelimits=false,
                ytick={-1, 0, ..., 6},
            ]

            \addplot[solid, color=blue]
            table{figsData/3rd_experiment/y_test.txt};
            \addlegendentry{Reference Solution}

            \addplot[name path=conf_mfkan_low,  fill=none, draw=none,forget plot]
            table{figsData/3rd_experiment/conf_mfkan_low.txt};
            \addplot[name path=conf_mfkan_high, fill=none, draw=none,forget plot]
            table{figsData/3rd_experiment/conf_mfkan_high.txt};

            \definecolor{mycolor1}{RGB}{75,200,75}
            \addplot[fill=mycolor1, fill opacity=0.85] fill between[of=conf_mfkan_high and conf_mfkan_low, ];
            \addlegendentry{Conformalized-MFKANs}

            \addplot[name path=mfkan_low,  fill=none, draw=none,forget plot]
            table{figsData/3rd_experiment/mfkan_low.txt};
            \addplot[name path=mfkan_high, fill=none, draw=none,forget plot]
            table{figsData/3rd_experiment/mfkan_high.txt};

            \addplot[fill=red, fill opacity=0.85] fill between[of=mfkan_high and mfkan_low, ];
            \addlegendentry{Ensemble MFKANs}

        \end{axis}
    \end{tikzpicture}

    \caption{Conformalized prediction intervals generated by ensemble MFKANs and Conformalized-MFKANs on the high-fidelity test dataset, computed with an ensemble size of 5 and a target miscoverage rate of \(\alpha = 0.05\).}
    \label{fig:MF_RESULTS}
\end{figure}
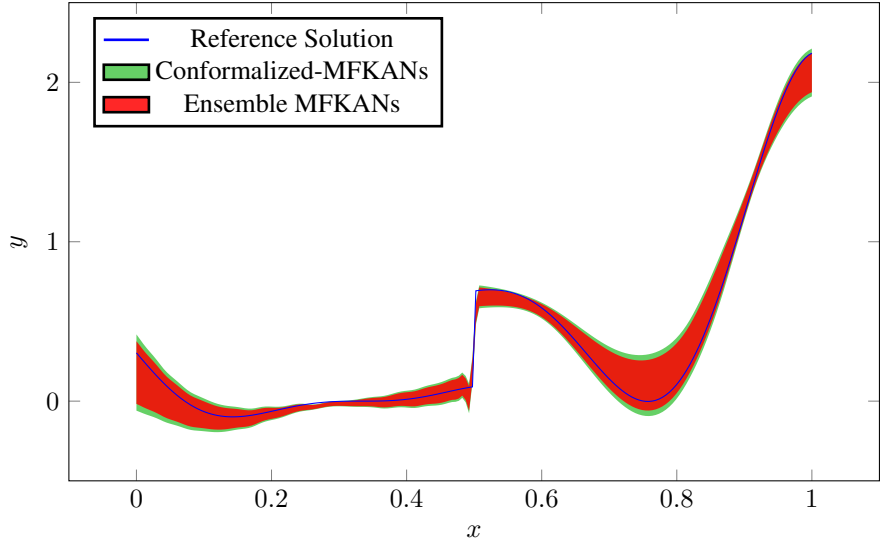

\textit{Results.} Figure \ref{fig:MF_RESULTS} illustrates the prediction intervals produced by the ensemble MFKANs, along with its conformalized predictions for the test dataset. Table \ref{table:MF} summarizes the performance metrics for each approach. Consistent with previous experiments, the Conformalized-MFKANs achieves the desired 95\% coverage, whereas the original, non-conformal model does not, indicating underestimation of uncertainty. Consequently, the intervals from the conformalized model are wider on average, reflecting a more reliable representation of uncertainty (see also the ablation study in Section \ref{app:ablation-mf}).

\begin{table}[t]
    \centering
    \begin{tabular}{c   c | c  c  }
        \hline
        \textbf{Model}      & \textbf{mean Coverage} & \textbf{Average PIW} & \textbf{Std.~Dev PIW} \\
        \hline
        Conformalized-MFKANs & \textbf{96.01}\%       & \textbf{0.18}        & \textbf{0.11}         \\
        Ensemble MFKANs              & 90.00\%                & 0.15                 & 0.09                  \\
        \hline
    \end{tabular}

        \caption{Average coverage, average interval width, and the standard deviation of interval widths for ensemble MFKANs and its conformalized counterpart, evaluated the high-fidelity test dataset. Each method targets a 95\% prediction interval, corresponding to a miscoverage rate of \(\alpha = 0.05\).}
    \label{table:MF}
\end{table}

\subsection{Experiment 4: PDE problem}
In this experiment, we analyze the wave equation, a second-order PDE, expressed as:

\[
    \frac{\partial^2 f}{\partial t^2} - c^2 \frac{\partial^2 f}{\partial x^2} = 0, \quad (x,t) \in [0,1] \times [0,1],
\]

subject to the following boundary and initial conditions:

\[
    f(0,t) = 0, \quad f(1,t) = 0, \quad t \in [0,1],
\]

\[
    f(x,0) = \sin(\pi x) + 0.5\sin(4\pi x), \quad f_t(x,0) = 0, \quad x \in [0,1].
\]

The exact solution is given by:

\[
    f(x,t) = \sin(\pi x) \cos(c\pi t) + 0.5 \sin(4\pi x) \cos(4c\pi t).
\]

We examine the case where \( c = \sqrt{2} \). {\color{black}{To train the KAN-based models, we employ a physics-informed learning framework in which we enforce supervision through initial and boundary condition data, along with physical consistency via the PDE residual. The total loss function consists of four components: the initial condition loss \( \mathcal{L}_{\text{IC}} \), which enforces consistency with the prescribed initial profile; the temporal derivative loss \( \mathcal{L}_{t} \), which imposes the initial velocity condition \( \frac{\partial f}{\partial t}(x,0) = 0 \); the boundary condition loss \( \mathcal{L}_{\text{BC}} \); and the residual loss \( \mathcal{L}_{\text{res}} \), which quantifies the violation of the governing PDE.

To prevent the residual term from dominating the optimization process and to ensure balanced contributions from all components, we apply a weighting factor \( \lambda_{\text{res}} = 0.01 \) to the residual loss. The resulting total loss function is defined as

\[
    \mathcal{L}_{\text{total}} = \mathcal{L}_{\text{IC}} + \mathcal{L}_{t} + \mathcal{L}_{\text{BC}} + \lambda_{\text{res}} \cdot \mathcal{L}_{\text{res}}.
\]

This formulation ensures that the model adheres to both empirical observations and the underlying physical laws, while maintaining numerical stability throughout training.}}

\textit{KAN Design and Datasets.} In this experiment, an ensemble of size 10 is utilized for both KAN and FBKAN models, with the number of subdomains for FBKANs set to 4. The dataset includes 1000 samples for training, 1200 for calibration, and 10000 for evaluation.

\begin{figure*}[h!]
    \centering
    \includegraphics[width=1\textwidth]{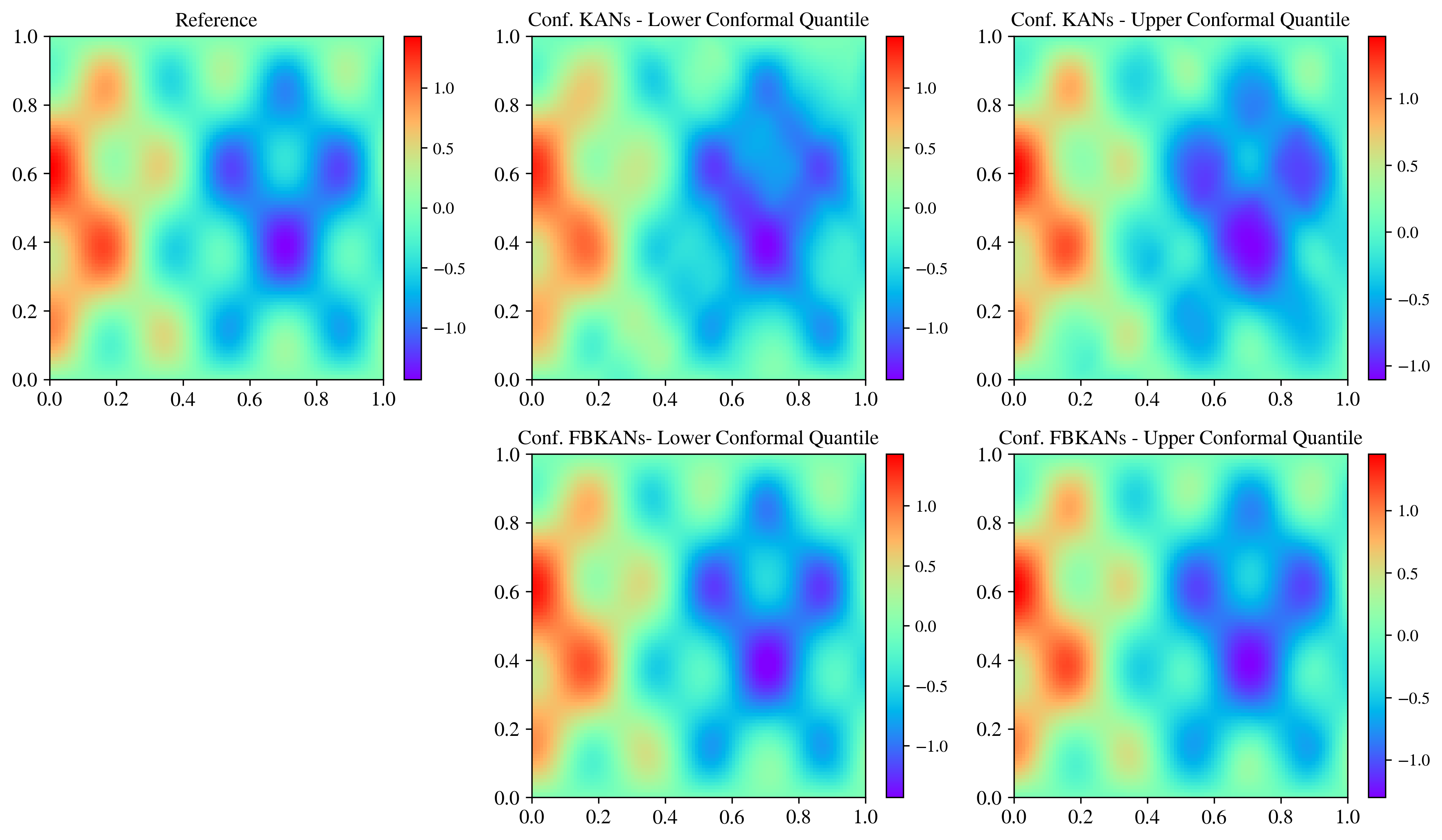}
    \caption{Conformalized prediction intervals generated by Conformalized-KANs and Conformalized-FBKANs on the wave equation test dataset, computed with an ensemble size of 10 and a target miscoverage rate of \(\alpha = 0.05\).}
    \label{fig:PDE-example}
\end{figure*}

\textit{Results.} Figure \ref{fig:PDE-example} depicts the conformalized prediction intervals produced by the KAN and FBKAN models, \textcolor{black}{along with the reference solution derived from the exact analytical expression of the wave equation across the spatiotemporal domain.} Table \ref{table:PDE-example} summarizes the corresponding coverage and width metrics. The results indicate that the ensemble models tend to underestimate uncertainty by producing narrower prediction intervals. In contrast, the conformalized versions correct this issue by widening the intervals appropriately, thereby achieving the target coverage of 95\% (see also the ablation study in Section \ref{app:ablation-pde}).

\begin{table}[h!]
    \centering
    \begin{tabular}{c   c   | c  c }
        \hline
        \textbf{Model}      & \textbf{mean Coverage} & \textbf{Average PIW} & \textbf{Std.~Dev PIW} \\

        \hline
        Conformalized-KANs   & \textbf{95.44}\%       & \textbf{0.20}        & \textbf{0.08}\        \\
        Conformalized-FBKANs & \textbf{95.34}\%       & \textbf{0.06}        & \textbf{0.03}\        \\
        \hline
        Ensemble KANs                 & 28.54\%                & 0.06                 & 0.02                  \\
        Ensemble FBKANs              & 25.80\%                & 0.01                 & 0.007                 \\
        \hline
    \end{tabular}
    \caption{Average coverage, average width, and standard deviation of widths for ensemble KANs, ensemble FBKANs, and their conformalized versions, evaluated on the wave equation test dataset. Each method targets a 95\% prediction interval, corresponding to a miscoverage rate of \(\alpha = 0.05\)}
   
    \label{table:PDE-example}
\end{table}

\section{Conclusions} \label{sec:conclusion}

\textcolor{black}{In this paper, we introduced Conformalized-KANs, a framework for UQ applied to KANs and their extensions, including FBKANs and MFKANs. By combining ensemble-based uncertainty estimation with conformal prediction, this approach generates prediction intervals with guaranteed coverage, without relying on restrictive distributional assumptions.}  Extensive numerical experiments across 1-D, 2-D, multi-fidelity, and PDE problems consistently demonstrated the effectiveness of Conformalized-KANs in guaranteeing the desired coverage of prediction intervals. Consequently, our results confirm that this framework significantly enhances the robustness and applicability of KAN models, addressing the crucial need for reliable UQ in data-limited scientific machine learning applications.

A notable outcome of our study is the consistent advantage of FBKANs over standard KANs in generating sharper prediction intervals. Even in their conformalized form, FBKANs produce significantly narrower intervals while still achieving the desired coverage guarantees, consistently outperforming KANs across all evaluated tasks. This underscores the strength of localized modeling via domain decomposition, which enables FBKANs to more accurately capture underlying function behavior and reduce epistemic uncertainty. Future work could build on these findings by incorporating adaptive subdomain partitioning strategies, allowing the model to dynamically adjust subdomain boundaries based on local uncertainty. This could potentially result in even sharper and better-calibrated prediction intervals. In addition, exploring advanced conformal prediction techniques, such as adaptive calibration sets, could enhance both the flexibility and efficiency of UQ in Conformalized-KANs.

\section*{Acknowledgments}
This project was completed with support from the U.S. Department of Energy, Advanced Scientific Computing Research program, under the "Uncertainty Quantification for Multifidelity Operator Learning (MOLUcQ)" project (Project No. 81739).  Pacific Northwest National Laboratory (PNNL) is a multi-program national laboratory operated for the U.S. Department of Energy (DOE) by Battelle Memorial Institute under Contract No. DE-AC05-76RL01830.

\appendix
\section{Appendix: Ablation Studies}
\label{app:ablation}
{\color{black}

This appendix presents ablation studies for the experiments discussed in Section~\ref{sec:numerical-experiments}. We investigate the sensitivity of Conformalized-KANs and their variants (FBKANs and MFKANs) to key hyperparameters. These studies provide insight into the robustness and practical tuning of our approach across diverse learning tasks.

\subsection{Experiment 1: 1-D Function}\label{app:ablation-1d}

\textbf{\textit{The Effect of Ensemble Size:}} We examine the impact of varying the ensemble size (\(M \in \{3, 4, 5, 6, 7\}\)) on the predicted intervals generated by ensemble KANs, ensemble FBKANs, and their conformalized counterparts. Figure \ref{subfig:Ablation1-a} shows that both Conformalized-KANs and Conformalized-FBKANs consistently achieve coverage close to the target 95\%, whereas their ensemble versions underestimate uncertainty.

Figure \ref{subfig:Ablation1-b} illustrates that for Conformalized-KANs, increasing the ensemble size improves the accuracy of the conformalized networks, as reflected in the reduction of the mean and standard deviation of the PIW. Notably, with an ensemble size of seven, the width distributions for Conformalized-KANs outperform those of its ensemble counterpart. In contrast, as seen in Figure \ref{subfig:Ablation1-c}, the mean and standard deviation of PIW for FBKANs models remain stable as the ensemble size increases.

\textbf{\textit{The Effect of the Number of Subdomains:}} We analyzed the effect of varying the number of subdomains (\(L \in \{3, 6, 10, 15, 20\}\)) on the predicted intervals for ensemble FBKANs and its conformalized counterpart. Figure \ref{subfig:Ablation1-d} demonstrates that Conformalized-FBKANs consistently achieves the target 95\% coverage, regardless of the number of subdomains. In contrast, the ensemble version consistently underestimates uncertainty across all tested configurations. This highlights the robustness of the Conformalized-FBKANs approach in providing reliable prediction intervals, even as problem complexity increases with a larger number of subdomains.

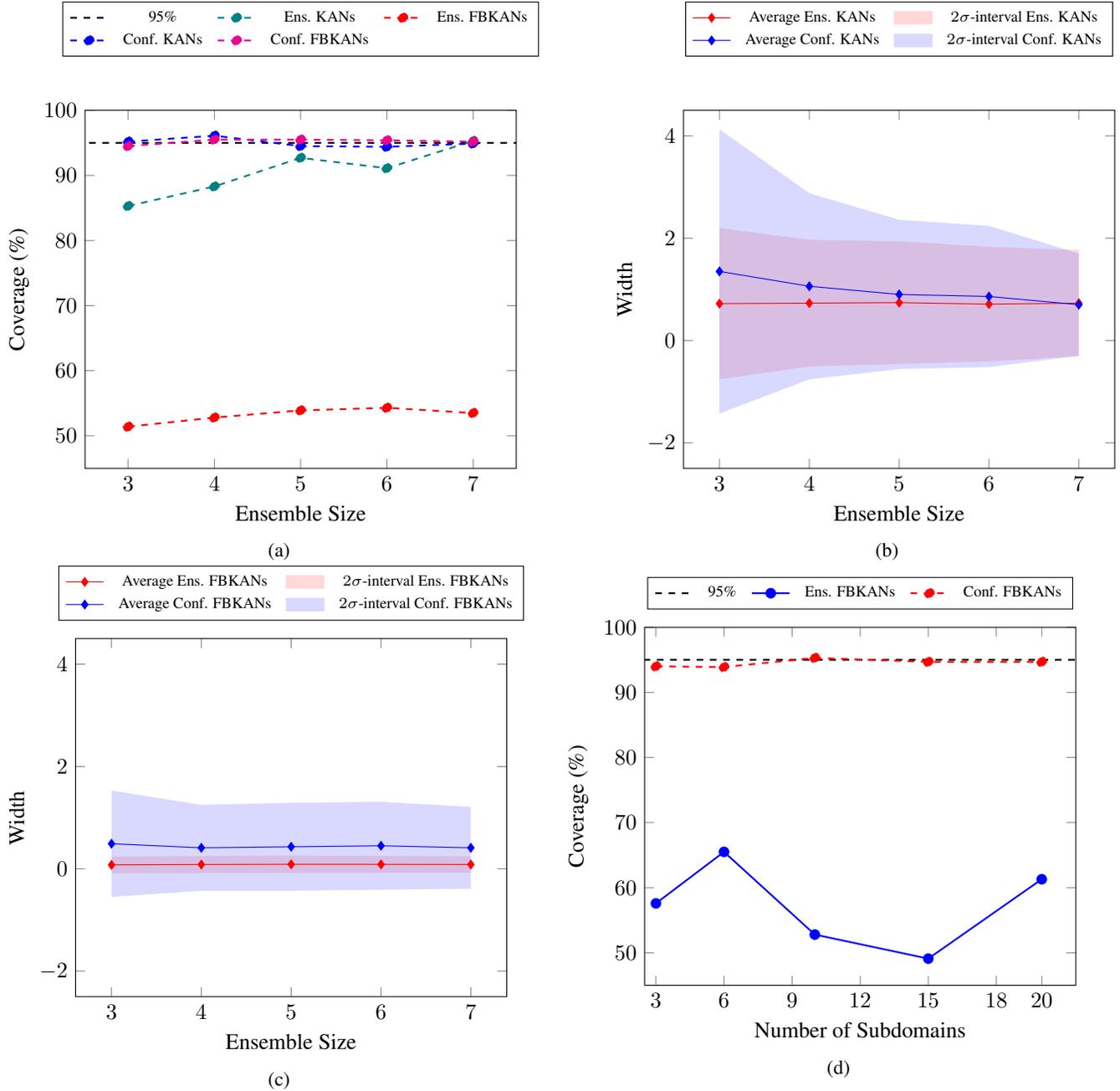
\begin{figure}[h!]
    \begin{subfigure}[!bl]{0.50\textwidth}
        \begin{tikzpicture}
            \begin{axis}[
                    xlabel={Ensemble Size}, 
                    ylabel={Coverage (\%)},
                    xmin=2.5, xmax=7.5,
                    ymin=45, ymax=100,
                    xtick={ 3,4,5,6,7},
                    ytick={50, 60,70,...,90,100},
                    legend style={
                            at={(0.5,1.15)}, 
                            anchor=south,
                            legend columns=3,
                            column sep=1ex
                        }
                ]
                \addplot [dashed, black, thick] coordinates {(0.7,95) (7.5,95)};
                \addlegendentry[font=\scriptsize]{95\%}
                \addplot [dashed, teal, mark=*, mark size=2pt, thick] table{figsData/1st_experiment/ens_KAN.txt};
                \addlegendentry[font=\scriptsize]{Ens. KANs}

                \addplot [dashed, red, mark=*, mark size=2pt, thick] table{figsData/1st_experiment/ens_fbKAN.txt};
                \addlegendentry[font=\scriptsize]{Ens. FBKANs}

                \addplot [dashed, blue, mark=*, mark size=2pt, thick] table{figsData/1st_experiment/ens_conf_KAN.txt};
                \addlegendentry[font=\scriptsize]{Conf. KANs}

                \addplot [dashed, magenta, mark=*, mark size=2pt, thick] table{figsData/1st_experiment/ens_conf_fbKAN.txt};
                \addlegendentry[font=\scriptsize]{Conf. FBKANs}

            \end{axis}
        \end{tikzpicture}
        \caption{}
        \label{subfig:Ablation1-a}
    \end{subfigure}
    \hspace{20pt}
    \begin{subfigure}[!br]{0.5\textwidth}
        \begin{tikzpicture}
            \begin{axis}[
                    xlabel={Ensemble Size},
                    ylabel={Width},
                    ymin=-2.5, ymax=4.5,
                    legend style={
                            at={(0.5,1.15)},
                            anchor=south,
                            legend columns=2,
                            column sep=1ex
                        }]


                \addplot[red, mark=diamond*, mark options={color=red}]
                coordinates {
                        (3,0.72)
                        (4,0.73)
                        (5,0.74)
                        (6,0.71)
                        (7,0.73)
                    };\addlegendentry[font=\scriptsize]{Average Ens. KANs}

                \addplot[name path=upper2, fill= none,draw=none,forget plot]
                coordinates {
                        (3,0.72+2*0.74)
                        (4,0.73+2*0.62)
                        (5,0.74+2*0.60)
                        (6,0.71+2*0.56)
                        (7,0.73+2*0.52)
                    };
                \addplot[name path=lower2, fill= none,draw=none,forget plot]
                coordinates {
                        (3,0.72-2*0.74)
                        (4,0.73-2*0.62)
                        (5,0.74-2*0.60)
                        (6,0.71-2*0.56)
                        (7,0.73-2*0.52)
                    };
                \addplot[red!50, opacity=0.3] fill between[of=upper2 and lower2];\addlegendentry[font=\scriptsize]{$2\sigma$-interval Ens. KANs}

                \addplot[blue, mark=diamond*, mark options={color=blue}]
                coordinates {
                        (3,1.35)
                        (4,1.06)
                        (5,0.90)
                        (6,0.86)
                        (7,0.70)
                    };\addlegendentry[font=\scriptsize]{Average Conf. KANs}

                \addplot[name path=upper1, fill= none,draw=none,forget plot]
                coordinates {
                        (3,1.35+2*1.39)
                        (4,1.06+2*0.91)
                        (5,0.90+2*0.73)
                        (6,0.86+2*0.69)
                        (7,0.70+2*0.50)
                    };
                \addplot[name path=lower1,fill= none, draw=none,forget plot]
                coordinates {
                        (3,1.35-2*1.39)
                        (4,1.06-2*0.91)
                        (5,0.90-2*0.73)
                        (6,0.86-2*0.69)
                        (7,0.70-2*0.50)
                    };
                \addplot[blue!50, opacity=0.3] fill between[of=upper1 and lower1,];\addlegendentry[font=\scriptsize]{$2\sigma$-interval Conf. KANs}

            \end{axis}

        \end{tikzpicture}
        \caption{}
        \label{subfig:Ablation1-b}

    \end{subfigure}
    \hspace{70pt}
    \begin{subfigure}[!br]{0.5\textwidth}
        \begin{tikzpicture}
            \begin{axis}[
                    xlabel={Ensemble Size},
                    ylabel={Width},
                    ymin=-2.5, ymax=4.5,
                    legend style={
                            at={(0.5,1.05)},
                            anchor=south,
                            legend columns=2,
                            column sep=1ex
                        }]


                \addplot[red, mark=diamond*, mark options={color=red}]
                coordinates {
                        (3,0.076)
                        (4,0.083)
                        (5,0.087)
                        (6,0.086)
                        (7,0.083)
                    };\addlegendentry[font=\scriptsize]{Average Ens. FBKANs}

                \addplot[name path=upper2, fill= none,draw=none,forget plot]
                coordinates {
                        (3,0.076+2*0.081)
                        (4,0.083+2*0.084)
                        (5,0.087+2*0.086)
                        (6,0.086+2*0.083)
                        (7,0.083+2*0.081)
                    };
                \addplot[name path=lower2, fill= none,draw=none,forget plot]
                coordinates {
                        (3,0.076-2*0.081)
                        (4,0.083-2*0.084)
                        (5,0.087-2*0.086)
                        (6,0.086-2*0.083)
                        (7,0.083-2*0.081)
                    };
                \addplot[red!50, opacity=0.3] fill between[of=upper2 and lower2];\addlegendentry[font=\scriptsize]{$2\sigma$-interval Ens. FBKANs}

                \addplot[blue, mark=diamond*, mark options={color=blue}]
                coordinates {
                        (3,0.49)
                        (4,0.41)
                        (5,0.43)
                        (6,0.45)
                        (7,0.41)
                    };\addlegendentry[font=\scriptsize]{Average Conf. FBKANs}

                \addplot[name path=upper1, fill= none,draw=none,forget plot]
                coordinates {
                        (3,0.49+2*0.52)
                        (4,0.41+2*0.42)
                        (5,0.43+2*0.43)
                        (6,0.45+2*0.43)
                        (7,0.41+2*0.40)
                    };
                \addplot[name path=lower1,fill= none, draw=none,forget plot]
                coordinates {
                        (3,0.49-2*0.52)
                        (4,0.41-2*0.42)
                        (5,0.43-2*0.43)
                        (6,0.45-2*0.43)
                        (7,0.41-2*0.40)
                    };
                \addplot[blue!50, opacity=0.3] fill between[of=upper1 and lower1,];\addlegendentry[font=\scriptsize]{$2\sigma$-interval Conf. FBKANs}

            \end{axis}

        \end{tikzpicture}
        \caption{}
        \label{subfig:Ablation1-c}

    \end{subfigure}
    \begin{subfigure}[!br]{0.5\textwidth}
        \begin{tikzpicture}
            \begin{axis}[
                    xlabel={Number of Subdomains}, 
                    ylabel={Coverage (\%)},
                    xmin=2.5, xmax=21.5,
                    ymin=45, ymax=100,
                    xtick={ 3,6,9,12,15,18,20},
                    ytick={50, 60,70,...,90,100},
                    legend style={
                            at={(0.5,1.05)}, 
                            anchor=south,
                            legend columns=3,
                            column sep=1ex
                        }
                ]
                \addplot [dashed, black, thick] coordinates {(0.7,95) (21.5,95)};
                \addlegendentry[font=\scriptsize]{95\%}

                \addplot [solid, blue, mark=*, mark size=2pt, thick, ] table{figsData/1st_experiment/domain_fbKAN.txt};
                \addlegendentry[font=\scriptsize]{Ens. FBKANs}

                \addplot [dashed, red, mark=*, mark size=2pt, thick] table{figsData/1st_experiment/domain_conf_fbKAN.txt};
                \addlegendentry[font=\scriptsize]{Conf. FBKANs}

            \end{axis}

        \end{tikzpicture}
        \caption{}
        \label{subfig:Ablation1-d}

    \end{subfigure}
    \begin{center}

    \end{center}
    \caption{Experiment 1: 1-D Function. (a) Coverage performance of ensemble KANs and ensemble FBKANs, along with their conformalized counterparts, across different ensemble sizes (\(M \in \{3, 4, 5, 6, 7\}\)). (b) Width distributions for conformalized and ensemble KANs as a function of ensemble size. (c) Width distributions for conformalized and ensemble FBKANs as a function of ensemble size. (d) Coverage performance of ensemble FBKANs and its conformalized version across different subdomain counts (\(L \in \{3, 6, 10, 15, 20\}\)).}
    \label{fig:Ablation1}
\end{figure}

\subsection{Experiment 3: Multi-Fidelity Problem}
\label{app:ablation-mf}
\textbf{\textit{The Effect of the Ensemble Size:}} Figure \ref{subfig:Ablation3-a} illustrates the effect of ensemble size (\(M \in \{3, 4, 5, 6, 7\}\)) on the prediction intervals of ensemble MFKAN and its conformalized counterpart. As shown, Conformalized-MFKANs maintains coverage close to the target 95\%, whereas its ensemble version falls short. However, as depicted in Figure \ref{subfig:Ablation3-b}, achieving the desired coverage with conformalization comes at the cost of wider prediction intervals.

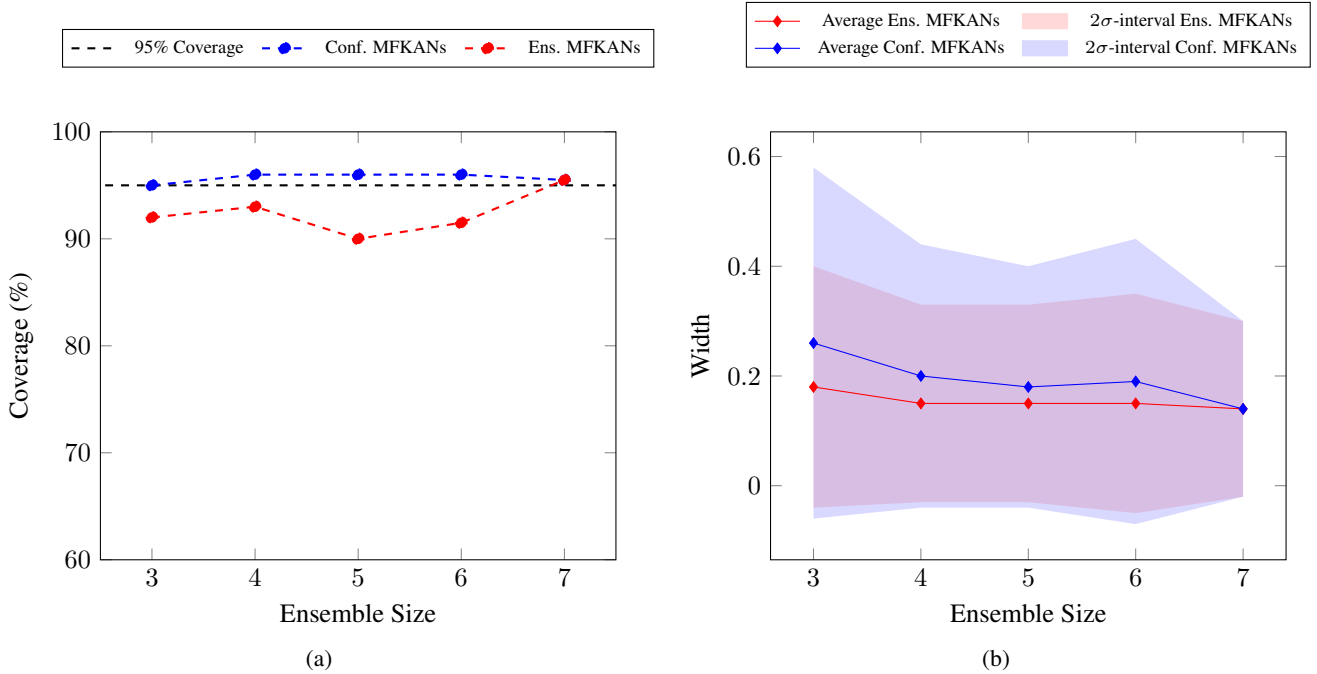
\begin{figure}[t!]
    \centering
    \begin{subfigure}[t]{0.48\textwidth}
        \centering
        \begin{tikzpicture}
            \begin{axis}[
                    xlabel={Ensemble Size},
                    ylabel={Coverage (\%)},
                    xmin=2.5, xmax=7.5,
                    ymin=60, ymax=100,
                    xtick={3,4,...,21},
                    ytick={60,70,...,90,100},
                    legend style={
                            at={(0.5,1.15)},
                            anchor=south,
                            legend columns=3,
                            column sep=1ex
                        }
                ]
                \addplot [dashed, black, thick] coordinates {(0.7,95) (21.5,95)};
                \addlegendentry[font=\scriptsize]{95\% Coverage }

                \addplot [dashed, blue, mark=*, mark size=2pt, thick] table{figsData/3rd_experiment/ens_conf_mfKAN.txt};
                \addlegendentry[font=\scriptsize]{Conf. MFKANs}

                \addplot [dashed, red, mark=*, mark size=2pt, thick] table{figsData/3rd_experiment/ens_mfKAN.txt};
                \addlegendentry[font=\scriptsize]{Ens. MFKANs}
            \end{axis}

        \end{tikzpicture}
        \caption{}
        \label{subfig:Ablation3-a}
    \end{subfigure}
    \hspace{10pt}
    \begin{subfigure}[t]{0.48\textwidth}
        \centering
        \begin{tikzpicture}
            \begin{axis}[
                    xlabel={Ensemble Size},
                    ylabel={Width},
                    legend style={
                            at={(0.5,1.15)},
                            anchor=south,
                            legend columns=2,
                            column sep=1ex
                        }
                ]
                \addplot[red, mark=diamond*, mark options={color=red}]
                coordinates {
                        (3,0.18)
                        (4,0.15)
                        (5,0.15)
                        (6,0.15)
                        (7,0.14)
                    };
                \addlegendentry[font=\scriptsize]{Average Ens. MFKANs}

                \addplot[name path=upper2, draw=none, forget plot]
                coordinates {
                        (3,0.18+2*0.11)
                        (4,0.15+2*0.09)
                        (5,0.15+2*0.09)
                        (6,0.15+2*0.10)
                        (7,0.14+2*0.08)
                    };
                \addplot[name path=lower2, draw=none, forget plot]
                coordinates {
                        (3,0.18-2*0.11)
                        (4,0.15-2*0.09)
                        (5,0.15-2*0.09)
                        (6,0.15-2*0.10)
                        (7,0.14-2*0.08)
                    };
                \addplot[red!50, opacity=0.3] fill between[of=upper2 and lower2];\addlegendentry[font=\scriptsize]{$2\sigma$-interval Ens. MFKANs}

                \addplot[blue, mark=diamond*, mark options={color=blue}]
                coordinates {
                        (3,0.26)
                        (4,0.20)
                        (5,0.18)
                        (6,0.19)
                        (7,0.14)
                    };
                \addlegendentry[font=\scriptsize]{Average Conf. MFKANs}

                \addplot[name path=upper1, draw=none, forget plot]
                coordinates {
                        (3,0.26+2*0.16)
                        (4,0.20+2*0.12)
                        (5,0.18+2*0.11)
                        (6,0.19+2*0.13)
                        (7,0.14+2*0.08)
                    };
                \addplot[name path=lower1, draw=none, forget plot]
                coordinates {
                        (3,0.26-2*0.16)
                        (4,0.20-2*0.12)
                        (5,0.18-2*0.11)
                        (6,0.19-2*0.13)
                        (7,0.14-2*0.08)
                    };
                \addplot[blue!50, opacity=0.3] fill between[of=upper1 and lower1];\addlegendentry[font=\scriptsize]{$2\sigma$-interval Conf. MFKANs}
            \end{axis}
        \end{tikzpicture}
        \caption{}
        \label{subfig:Ablation3-b}
    \end{subfigure}

    \caption{Experiment 3: Multi-Fidelity Problem. (a) Coverage performance of ensemble MFKANs, along with its conformalized version, across different ensemble sizes (\(M \in \{3, 4, 5, 6, 7\}\)). (b) Width distributions for ensemble and conformalized MFKANs as a function of ensemble size.}
    \label{fig:MF}
\end{figure}

\subsection{Experiment 4: PDE problem}
\label{app:ablation-pde}

\textbf{\textit{The Effect of the Ensemble Size:}} Figure \ref{subfig:Ablation4-a} illustrates the impact of varying the ensemble size (\(M \in \{3, 4, 5, 6, 7, 8, 9, 10\}\)) on the predicted intervals for ensemble KANs, ensemble FBKANs, and their conformalized counterparts. The results show that ensemble models yield lower coverage, resulting in underestimated uncertainty. In contrast, the conformalized models refine the predictions, maintaining coverage near the target 95\% level.

\textbf{\textit{The Effect of the Size of the Calibration Datasets:}} Figure \ref{subfig:Ablation4-b} illustrates the impact of calibration dataset size on the coverage probability of conformalized-KANs and conformalized-FBKANs for an ensemble of 5. As shown in the figure, smaller calibration datasets lead to fluctuations in coverage, deviating from the target 95\% level. However, as the dataset size increases, these variations diminish, resulting in more stable coverage that aligns with the desired confidence level. This highlights the crucial role of adequate calibration data in ensuring reliable prediction confidence.

\begin{figure}[t!]
    \begin{subfigure}[!bl]{0.48\textwidth}
        \begin{tikzpicture}
            \begin{axis}[
                    xlabel={Ensemble Size}, 
                    ylabel={Coverage (\%)},
                    xmin=2.5, xmax=10.5,
                    ymin=10, ymax=105,
                    xtick={ 3,4,...,9,10},
                    ytick={10,20,...,90,100},
                    legend style={
                            at={(0.5,1.05)}, 
                            anchor=south,
                            legend columns=3,
                            column sep=1ex
                        }
                ]
                \addplot [dashed, black, thick] coordinates {(0.7,95) (10.5,95)};
                \addlegendentry[font=\scriptsize]{95\%}
                \addplot [dashed, teal, mark=*, mark size=2pt, thick] table{figsData/4th_experiment/ens_KAN.txt};
                \addlegendentry[font=\scriptsize]{Ens. KANs}

                \addplot [dashed, red, mark=*, mark size=2pt, thick] table{figsData/4th_experiment/ens_fbKAN.txt};
                \addlegendentry[font=\scriptsize]{Ens. FBKANs}

                \addplot [dashed, blue, mark=*, mark size=2pt, thick] table{figsData/4th_experiment/ens_conf_KAN.txt};
                \addlegendentry[font=\scriptsize]{Conf. KANs}

                \addplot [dashed, magenta, mark=*, mark size=2pt, thick] table{figsData/4th_experiment/ens_conf_fbKAN.txt};
                \addlegendentry[font=\scriptsize]{Conf. FBKANs}

            \end{axis}
        \end{tikzpicture}
        
        \caption{}
        \label{subfig:Ablation4-a}

    \end{subfigure}
    \hspace{5pt}
    \begin{subfigure}[!br]{0.48\textwidth}
        \begin{tikzpicture}
            \begin{axis}[
                    xlabel={Calibration Datasets Size}, 
                    ylabel={Coverage (\%)},
                    xmin=15, xmax=2015,
                    ymin=87, ymax=102,
                    xtick={ 20,50,...,1000},
                    xtick={20,300,600,1000,1500,2000},
                    ytick={90,95,100},
                    legend style={
                            at={(0.5,1.05)}, 
                            anchor=south,
                            legend columns=3,
                            column sep=1ex
                        }
                ]
                \addplot [dashed, black, thick] coordinates {(15,95) (2015.5,95)};
                \addlegendentry[font=\scriptsize]{95\% Coverage }

                \addplot [dashed, blue, mark=*, mark size=2pt, thick] table{figsData/4th_experiment/cal_conf_KAN.txt};
                \addlegendentry[font=\scriptsize]{Conf. KANs}


                \addplot [dashed, red, mark=*, mark size=2pt, thick] table{figsData/4th_experiment/cal_conf_fbKAN.txt};
                \addlegendentry[font=\scriptsize]{Conf. FBKANs}

            \end{axis}

        \end{tikzpicture}
        \caption{}
        \label{subfig:Ablation4-b}

    \end{subfigure}

    \caption{Experiment 4: PDE Problem. (a) Coverage performance of ensemble KANs and ensemble FBKANs, along with their conformalized counterparts, across different ensemble sizes (\(M \in \{3, 4, 5, 6, 7, 8, 9, 10\}\)).  (b) Coverage performance of Conformalized-KANs and Conformalized-FBKANs with respect to calibration dataset size, evaluated using an ensemble of size 5.}

    \label{fig:Ablation3}
\end{figure}
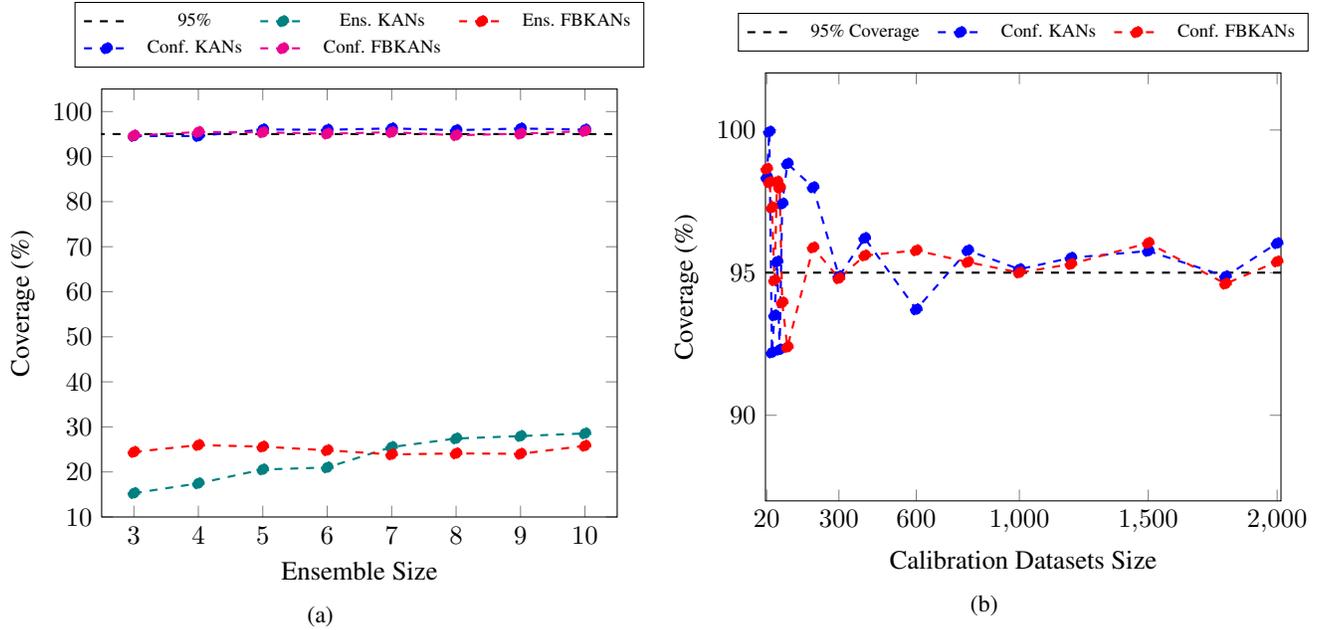

}
\bibliographystyle{unsrt}
\bibliography{references.bib}

\begin{thebibliography}{10}

\bibitem{baker_workshop_2019}
Nathan Baker, Frank Alexander, Timo Bremer, Aric Hagberg, Yannis Kevrekidis, Habib Najm, Manish Parashar, Abani Patra, James Sethian, Stefan Wild, Karen Willcox, and Steven Lee.
\newblock Workshop {Report} on {Basic} {Research} {Needs} for {Scientific} {Machine} {Learning}: {Core} {Technologies} for {Artificial} {Intelligence}.
\newblock Technical report, USDOE Office of Science (SC), Washington, D.C. (United States), February 2019.

\bibitem{karniadakis2021physics}
George~Em Karniadakis, Ioannis~G Kevrekidis, Lu~Lu, Paris Perdikaris, Sifan Wang, and Liu Yang.
\newblock Physics-informed machine learning.
\newblock {\em Nature Reviews Physics}, 3(6):422--440, 2021.

\bibitem{liu2024kan}
Ziming Liu, Yixuan Wang, Sachin Vaidya, Fabian Ruehle, James Halverson, Marin Solja{\v{c}}i{\'c}, Thomas~Y Hou, and Max Tegmark.
\newblock Kan: Kolmogorov-arnold networks.
\newblock {\em arXiv preprint arXiv:2404.19756}, 2024.

\bibitem{liu2024kan2}
Ziming Liu, Pingchuan Ma, Yixuan Wang, Wojciech Matusik, and Max Tegmark.
\newblock Kan 2.0: Kolmogorov-arnold networks meet science.
\newblock {\em arXiv preprint arXiv:2408.10205}, 2024.

\bibitem{kolmogorov1957representations}
Andrei~Nikolaevich Kolmogorov.
\newblock On the representations of continuous functions of many variables by superposition of continuous functions of one variable and addition.
\newblock In {\em Dokl. Akad. Nauk USSR}, volume 114, pages 953--956, 1957.

\bibitem{yu2024kan}
Runpeng Yu, Weihao Yu, and Xinchao Wang.
\newblock Kan or mlp: A fairer comparison.
\newblock {\em arXiv preprint arXiv:2407.16674}, 2024.

\bibitem{zeng2024kan}
Chen Zeng, Jiahui Wang, Haoran Shen, and Qiao Wang.
\newblock Kan versus mlp on irregular or noisy functions.
\newblock {\em arXiv preprint arXiv:2408.07906}, 2024.

\bibitem{shukla2024comprehensive}
Khemraj Shukla, Juan~Diego Toscano, Zhicheng Wang, Zongren Zou, and George~Em Karniadakis.
\newblock A comprehensive and fair comparison between mlp and kan representations for differential equations and operator networks.
\newblock {\em arXiv preprint arXiv:2406.02917}, 2024.

\bibitem{wang2024kolmogorov}
Yizheng Wang, Jia Sun, Jinshuai Bai, Cosmin Anitescu, Mohammad~Sadegh Eshaghi, Xiaoying Zhuang, Timon Rabczuk, and Yinghua Liu.
\newblock Kolmogorov arnold informed neural network: A physics-informed deep learning framework for solving pdes based on kolmogorov arnold networks.
\newblock {\em arXiv preprint arXiv:2406.11045}, 2024.

\bibitem{rigas2024adaptive}
Spyros Rigas, Michalis Papachristou, Theofilos Papadopoulos, Fotios Anagnostopoulos, and Georgios Alexandridis.
\newblock Adaptive training of grid-dependent physics-informed kolmogorov-arnold networks.
\newblock {\em IEEE Access}, 2024.

\bibitem{patra2024physics}
Subhajit Patra, Sonali Panda, Bikram~Keshari Parida, Mahima Arya, Kurt Jacobs, Denys~I Bondar, and Abhijit Sen.
\newblock Physics informed kolmogorov-arnold neural networks for dynamical analysis via efficent-kan and wav-kan.
\newblock {\em arXiv preprint arXiv:2407.18373}, 2024.

\bibitem{genet2024tkan}
Remi Genet and Hugo Inzirillo.
\newblock Tkan: Temporal kolmogorov-arnold networks.
\newblock {\em arXiv preprint arXiv:2405.07344}, 2024.

\bibitem{bozorgasl2024wav}
Zavareh Bozorgasl and Hao Chen.
\newblock Wav-kan: Wavelet kolmogorov-arnold networks.
\newblock {\em arXiv preprint arXiv:2405.12832}, 2024.

\bibitem{seydi2024unveiling}
Seyd~Teymoor Seydi, Zavareh Bozorgasl, and Hao Chen.
\newblock Unveiling the power of wavelets: A wavelet-based kolmogorov-arnold network for hyperspectral image classification.
\newblock {\em arXiv preprint arXiv:2406.07869}, 2024.

\bibitem{meshir2025study}
Juan~Daniel Meshir, Abel Palafox, and Edgar~Alejandro Guerrero.
\newblock On the study of frequency control and spectral bias in wavelet-based kolmogorov arnold networks: A path to physics-informed kans.
\newblock {\em arXiv preprint arXiv:2502.00280}, 2025.

\bibitem{kiamari2024gkan}
Mehrdad Kiamari, Mohammad Kiamari, and Bhaskar Krishnamachari.
\newblock Gkan: Graph kolmogorov-arnold networks.
\newblock {\em arXiv preprint arXiv:2406.06470}, 2024.

\bibitem{zhang2024graphkan}
Fan Zhang and Xin Zhang.
\newblock Graphkan: Enhancing feature extraction with graph kolmogorov arnold networks.
\newblock {\em arXiv preprint arXiv:2406.13597}, 2024.

\bibitem{de2024kolmogorov}
Gianluca De~Carlo, Andrea Mastropietro, and Aris Anagnostopoulos.
\newblock Kolmogorov-arnold graph neural networks.
\newblock {\em arXiv preprint arXiv:2406.18354}, 2024.

\bibitem{ss2024chebyshev}
Sidharth SS, Keerthana AR, Anas KP, et~al.
\newblock Chebyshev polynomial-based kolmogorov-arnold networks: An efficient architecture for nonlinear function approximation.
\newblock {\em arXiv preprint arXiv:2405.07200}, 2024.

\bibitem{mostajeran2025scaled}
Farinaz Mostajeran and Salah~A Faroughi.
\newblock Scaled-cpikans: Domain scaling in chebyshev-based physics-informed kolmogorov-arnold networks.
\newblock {\em arXiv preprint arXiv:2501.02762}, 2025.

\bibitem{aghaei2024fkan}
Alireza~Afzal Aghaei.
\newblock fkan: Fractional kolmogorov-arnold networks with trainable jacobi basis functions.
\newblock {\em arXiv preprint arXiv:2406.07456}, 2024.

\bibitem{abueidda2024deepokan}
Diab~W Abueidda, Panos Pantidis, and Mostafa~E Mobasher.
\newblock Deepokan: Deep operator network based on kolmogorov arnold networks for mechanics problems.
\newblock {\em arXiv preprint arXiv:2405.19143}, 2024.

\bibitem{vaca2024kolmogorov}
Cristian~J Vaca-Rubio, Luis Blanco, Roberto Pereira, and M{\`a}rius Caus.
\newblock Kolmogorov-arnold networks (kans) for time series analysis.
\newblock {\em arXiv preprint arXiv:2405.08790}, 2024.

\bibitem{kashefi2024kolmogorov}
Ali Kashefi.
\newblock Kolmogorov-arnold pointnet: Deep learning for prediction of fluid fields on irregular geometries.
\newblock {\em arXiv preprint arXiv:2408.02950}, 2024.

\bibitem{toscano2024inferring}
Juan~Diego Toscano, Theo K{\"a}ufer, Martin Maxey, Christian Cierpka, and George~Em Karniadakis.
\newblock Inferring turbulent velocity and temperature fields and their statistics from lagrangian velocity measurements using physics-informed kolmogorov-arnold networks.
\newblock {\em arXiv preprint arXiv:2407.15727}, 2024.

\bibitem{azam2024suitability}
Basim Azam and Naveed Akhtar.
\newblock Suitability of kans for computer vision: A preliminary investigation.
\newblock {\em arXiv preprint arXiv:2406.09087}, 2024.

\bibitem{cheon2024demonstrating}
Minjong Cheon.
\newblock Demonstrating the efficacy of kolmogorov-arnold networks in vision tasks.
\newblock {\em arXiv preprint arXiv:2406.14916}, 2024.

\bibitem{howard2024finite}
Amanda~A Howard, Bruno Jacob, Sarah~H Murphy, Alexander Heinlein, and Panos Stinis.
\newblock Finite basis kolmogorov-arnold networks: domain decomposition for data-driven and physics-informed problems.
\newblock {\em arXiv preprint arXiv:2406.19662}, 2024.

\bibitem{howard2024multifidelity}
Amanda~A Howard, Bruno Jacob, and Panos Stinis.
\newblock Multifidelity kolmogorov-arnold networks.
\newblock {\em arXiv preprint arXiv:2410.14764}, 2024.

\bibitem{raissi_physics-informed_2019}
M.~Raissi, P.~Perdikaris, and G.~E. Karniadakis.
\newblock Physics-informed neural networks: a deep learning framework for solving forward and inverse problems involving nonlinear partial differential equations.
\newblock {\em Journal of Computational Physics}, 378:686--707, 2019.

\bibitem{heinlein_multifidelity_2024}
Alexander Heinlein, Amanda~A. Howard, Damien Beecroft, and Panos Stinis.
\newblock Multifidelity domain decomposition-based physics-informed neural networks and operators for time-dependent problems, June 2024.
\newblock arXiv:2401.07888 [cs, math].

\bibitem{psaros2023uncertainty}
Apostolos~F Psaros, Xuhui Meng, Zongren Zou, Ling Guo, and George~Em Karniadakis.
\newblock Uncertainty quantification in scientific machine learning: Methods, metrics, and comparisons.
\newblock {\em Journal of Computational Physics}, 477:111902, 2023.

\bibitem{yang2022scalable}
Yibo Yang, Georgios Kissas, and Paris Perdikaris.
\newblock Scalable uncertainty quantification for deep operator networks using randomized priors.
\newblock {\em Computer Methods in Applied Mechanics and Engineering}, 399:115399, 2022.

\bibitem{lin2023b}
Guang Lin, Christian Moya, and Zecheng Zhang.
\newblock B-deeponet: An enhanced bayesian deeponet for solving noisy parametric pdes using accelerated replica exchange sgld.
\newblock {\em Journal of Computational Physics}, 473:111713, 2023.

\bibitem{moya2023deeponet}
Christian Moya, Shiqi Zhang, Guang Lin, and Meng Yue.
\newblock Deeponet-grid-uq: A trustworthy deep operator framework for predicting the power grid’s post-fault trajectories.
\newblock {\em Neurocomputing}, 535:166--182, 2023.

\bibitem{zhang2024bayesian}
Zecheng Zhang, Christian Moya, Wing~Tat Leung, Guang Lin, and Hayden Schaeffer.
\newblock Bayesian deep operator learning for homogenized to fine-scale maps for multiscale pde.
\newblock {\em Multiscale Modeling \& Simulation}, 22(3):956--972, 2024.

\bibitem{giroux2024uncertainty}
James Giroux and Cristiano Fanelli.
\newblock Uncertainty quantification with bayesian higher order relu-kans.
\newblock {\em Machine Learning: Science and Technology}, 2024.

\bibitem{hassan2024bayesian}
Masoud~Muhammed Hassan.
\newblock Bayesian kolmogorov arnold networks (bayesian\_kans): A probabilistic approach to enhance accuracy and interpretability.
\newblock {\em arXiv preprint arXiv:2408.02706}, 2024.

\bibitem{SAHIN2024124813}
Izzet Sahin, Christian Moya, Amirhossein Mollaali, Guang Lin, and Guillermo Paniagua.
\newblock Deep operator learning-based surrogate models with uncertainty quantification for optimizing internal cooling channel rib profiles.
\newblock {\em International Journal of Heat and Mass Transfer}, 219:124813, 2024.

\bibitem{vovk2005algorithmic}
Vladimir Vovk, Alexander Gammerman, and Glenn Shafer.
\newblock {\em Algorithmic learning in a random world}, volume~29.
\newblock Springer, 2005.

\bibitem{romano2019conformalized}
Yaniv Romano, Evan Patterson, and Emmanuel Candes.
\newblock Conformalized quantile regression.
\newblock {\em Advances in neural information processing systems}, 32, 2019.

\bibitem{angelopoulos2021gentle}
Anastasios~N Angelopoulos and Stephen Bates.
\newblock A gentle introduction to conformal prediction and distribution-free uncertainty quantification.
\newblock {\em arXiv preprint arXiv:2107.07511}, 2021.

\bibitem{MOYA2025134418}
Christian Moya, Amirhossein Mollaali, Zecheng Zhang, Lu~Lu, and Guang Lin.
\newblock Conformalized-deeponet: A distribution-free framework for uncertainty quantification in deep operator networks.
\newblock {\em Physica D: Nonlinear Phenomena}, 471:134418, 2025.

\bibitem{mollaali2024conformalized}
Amirhossein Mollaali, Gabriel Zufferey, Gonzalo Constante-Flores, Christian Moya, Can Li, Guang Lin, and Meng Yue.
\newblock Conformalized prediction of post-fault voltage trajectories using pre-trained and finetuned attention-driven neural operators.
\newblock {\em arXiv preprint arXiv:2410.24162}, 2024.

\bibitem{moseley2023finite}
Ben Moseley, Andrew Markham, and Tarje Nissen-Meyer.
\newblock Finite basis physics-informed neural networks (fbpinns): a scalable domain decomposition approach for solving differential equations.
\newblock {\em Advances in Computational Mathematics}, 49(4):62, 2023.

\bibitem{dolean2024multilevel}
Victorita Dolean, Alexander Heinlein, Siddhartha Mishra, and Ben Moseley.
\newblock Multilevel domain decomposition-based architectures for physics-informed neural networks.
\newblock {\em Computer Methods in Applied Mechanics and Engineering}, 429:117116, 2024.

\bibitem{anderson2024elm}
Samuel Anderson, Victorita Dolean, Ben Moseley, and Jennifer Pestana.
\newblock Elm-fbpinn: efficient finite-basis physics-informed neural networks.
\newblock {\em arXiv preprint arXiv:2409.01949}, 2024.

\bibitem{heinlein2024multifidelity}
Alexander Heinlein, Amanda~A Howard, Damien Beecroft, and Panos Stinis.
\newblock Multifidelity domain decomposition-based physics-informed neural networks for time-dependent problems.
\newblock {\em arXiv preprint arXiv:2401.07888}, 2024.

\bibitem{meng2020composite}
Xuhui Meng and George~Em Karniadakis.
\newblock A composite neural network that learns from multi-fidelity data: Application to function approximation and inverse pde problems.
\newblock {\em Journal of Computational Physics}, 401:109020, 2020.

\end{thebibliography}


\end{document}